\documentclass[10pt,twocolumn,letterpaper]{article}
\usepackage{cvpr}              
\makeatletter
\@namedef{ver@everyshi.sty}{}
\makeatother
\usepackage{graphicx}
\graphicspath{{figures/}}
\usepackage{amsmath}
\usepackage{amssymb}
\usepackage{booktabs}
\usepackage{algorithm}
\usepackage{algorithmicx}
\usepackage{algpseudocode}
\usepackage{multirow}
\usepackage[table,xcdraw]{xcolor}
\usepackage{booktabs}
\usepackage{verbatim}
\usepackage{color, colortbl}
\usepackage{bm}
\usepackage{enumitem}
\usepackage{pifont}
\usepackage[accsupp]{axessibility}
\usepackage{multirow}
\usepackage{makecell}
\usepackage{adjustbox}
\usepackage{bbding}
\usepackage{tikz}
\usepackage{pgfplots}
\usepackage{ulem}

\usepackage[pagebackref,breaklinks,colorlinks]{hyperref}

\usepackage[capitalize]{cleveref}
\crefname{section}{Sec.}{Secs.}
\Crefname{section}{Section}{Sections}
\Crefname{table}{Table}{Tables}
\crefname{table}{Tab.}{Tabs.}

\newcommand{\cmark}{\ding{51}}%
\newcommand{\xmark}{\ding{55}}%

\definecolor{Gray}{gray}{0.9}
\newcommand{\best}[1]{\color{red}\textbf{#1}}
\newcommand{\second}[1]{\color{blue}\textbf{#1}}

\def\cvprPaperID{32} 
\def\confName{CVPR}
\def\confYear{2022}
\newcommand\blfootnote[1]{%
  \begingroup
  \renewcommand\thefootnote{}\footnote{#1}%
  \addtocounter{footnote}{-1}%
  \endgroup
}

\pgfplotsset{compat=1.17}

\begin{document}

\title{CFA: Constraint-based Finetuning Approach \\ for Generalized Few-Shot Object Detection}

\author{{\normalsize Karim Guirguis}$^{1 \dagger}$ \hspace{0.1em} {\normalsize Ahmed Hendawy}$^{1,2 \dagger}$ \hspace{0.1em}  {\normalsize George Eskandar}$^{2}$ \hspace{0.1em}  {\normalsize Mohamed Abdelsamad}$^{1,2}$ \hspace{0.1em}  {\normalsize Matthias Kayser}$^{1}$ \hspace{0.1em}  {\normalsize J\"urgen Beyerer}$^{3,4}$\\
{\normalsize Robert Bosch GmbH}$^1$ \hspace{0.2em} {\normalsize University of Stuttgart}$^2$ \hspace{0.2em} {\normalsize Karlsruhe Institute of Technology}$^3$ \hspace{0.2em} {\normalsize Fraunhofer IOSB}$^4$ \\
{\tt\small karim.guirguis@de.bosch.com}
}
\maketitle

\begin{abstract}
Few-shot object detection (FSOD) seeks to detect novel categories with limited data by leveraging prior knowledge from abundant base data. Generalized few-shot object detection (G-FSOD) aims to tackle FSOD without forgetting previously seen base classes and, thus, accounts for a more realistic scenario, where both classes are encountered during test time. While current FSOD methods suffer from catastrophic forgetting, G-FSOD addresses this limitation yet exhibits a performance drop on novel tasks compared to the state-of-the-art FSOD. In this work, we propose a constraint-based finetuning approach (CFA) to alleviate catastrophic forgetting, while achieving competitive results on the novel task without increasing the model capacity. CFA adapts a continual learning method, namely Average Gradient Episodic Memory (A-GEM) to G-FSOD. Specifically, more constraints on the gradient search strategy are imposed from which a new gradient update rule is derived, allowing for better knowledge exchange between base and novel classes. To evaluate our method, we conduct extensive experiments on MS-COCO and PASCAL-VOC datasets. Our method outperforms current FSOD and G-FSOD approaches on the novel task with minor degeneration on the base task. Moreover, CFA is orthogonal to FSOD approaches and operates as a plug-and-play module without increasing the model capacity or inference time.
\end{abstract}
\blfootnote{$\dagger$ Authors have equally contributed to this work.}

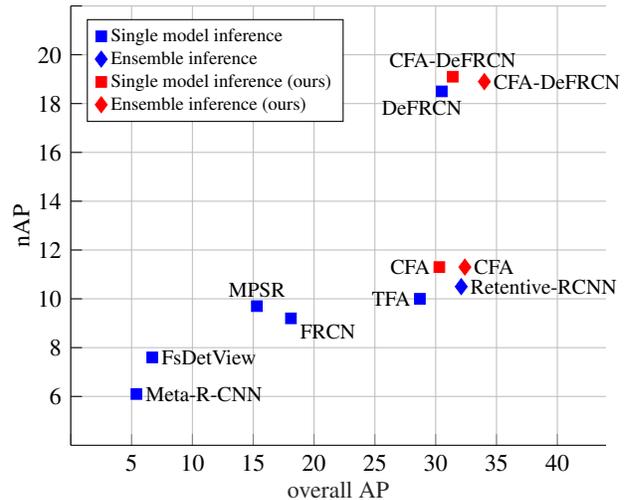
\begin{figure}[t!]
    \begin{tikzpicture}
\begin{axis}[%
width=2.8in,
height=2.3in,
scale only axis,
xticklabel style = {font=\small},
xmin=0,
xmax=44,
xlabel style={font=\color{white!15!black}, at={(axis description cs:0.5,-0.06)}, anchor=north},
xlabel={\small overall AP},
every outer y axis line/.append style={black},
every y tick label/.append style={font=\small \color{black}},
every y tick/.append style={black},
yticklabel style={%
	/pgf/number format/.cd,
	fixed,
	fixed zerofill,
	precision=3,
},
xtick={0,5,10,15,20,25,30, 35, 40, 45},
xticklabels = {{},{5},{10},{15},{20},{25},{30}, {35}, {40}, {45}},
ymin=4,
ymax=22,
ytick={0,6,8,10, 12, 14, 16, 18, 20, 22},
yticklabels = {{},{6},{8},{10},{12},{14},{16},{18},{20}, {}},
ylabel style={font=\color{black}, at={(axis description cs:-0.06,0.5)}, anchor=south},
ylabel={\small \textcolor{black}{nAP}},
axis background/.style={fill=white},
axis x line*=bottom,
axis y line*=left,
xmajorgrids,
ymajorgrids,
legend pos = north west,
legend style={legend cell align=left, align=left, draw=black, {nodes={scale=0.70}}}
]

\addplot [only marks,
scatter src=explicit symbolic,
scatter/classes={a={mark=square*, draw=blue, mark size=2pt, fill=blue}, b={mark=diamond*, draw=blue, fill=blue, mark size=3pt}, c={mark=square*, draw=red, mark size=2pt, fill=red}, d={mark=diamond*, draw=red, fill=red, mark size=3pt}},
nodes near coords*={\footnotesize \Label},
visualization depends on={value \thisrow{label} \as \Label},
visualization depends on=\thisrow{alignment} \as \alignment,
every node near coord/.style={anchor=\alignment}]
table[row sep=crcr, meta=class]{
x   y   label            class   alignment\\
28.7 10.0 TFA               a        0\\
30.3 11.3 CFA               c        0\\
31.4 19.1 CFA-DeFRCN        c        -90\\
32.1 10.5 Retentive-RCNN         b        180\\
30.5 18.5 DeFRCN            a        40 \\
32.4 11.3 CFA               d        -180\\
15.3 9.7  MPSR              a        -90\\
6.7 7.6  FsDetView          a        180\\
18.1 9.2 FRCN               a        160\\
5.4 6.1  Meta-R-CNN         a        180\\
34.0 18.9 CFA-DeFRCN        d        180\\
};




\legend{Single model inference, Ensemble inference, Single model inference (ours), Ensemble inference (ours)}

\end{axis}


\end{tikzpicture}%
    \caption{Performance of FSOD methods in G-FSOD task on MS-COCO dataset under 10-shot setting. The proposed finetuning approach, CFA, is applied to various FSOD frameworks, enhancing their performance in the overall and novel AP. Inference with an ensemble of base and novel models result in the highest overall AP but suffer from an increased in computations and inference time. Inference with a single CFA-finetuned model achieves the best trade-off between performance and speed.}
    \vspace{-2em}
    \label{fig:teaser}
\end{figure}

\section{Introduction}
\label{sec:intro}
Object detection is one fundamental building block for visual perception systems. Given an image, the class label and the spatial location are jointly identified for each object. In recent years, the rise of deep convolutional neural networks (CNN) \cite{CNN} has enabled a wide variety of data-driven object detectors\cite{R-CNN, FastR-CNN, FasterR-CNN, SSD, YOLOv1, YOLOv2, EfficientDet}, delivering competitive performance. However, the training of object detectors requires a significant amount of labeled images, which are time-consuming and labor-intensive to acquire and to label.

Inspired by the human cognitive ability of learning new concepts with few examples, the few-shot learning (FSL) paradigm has been introduced to tackle the aforementioned issues. This is achieved by leveraging prior knowledge from previous tasks with abundant data to rapidly learn new concepts under a low-data regime. Currently, there are three different ways to integrate this prior knowledge \cite{GFSL}: (1) learning a data augmentation to increase the number of data samples, (2) creating a model that limits the hypothesis space, or (3) developing an algorithm that enables an effective exploration of the hypothesis space.  

In 2018, Chen et al. \cite{LSTD} introduced the first few-shot object detection (FSOD) framework. The aim of FSOD is to adapt object detectors to learn novel categories with limited data. Since  then,  a  number  of  progressively  enhanced  frameworks were introduced, which can be grouped into two main approaches: meta-learning based and transfer learning based methods. Meta-learning based methods \cite{RepMet, FSRW, MetaDet, MetaRCNN, FSOD-RPN, FsDetView, CME} perform an instance-level exemplar search utilizing a support set of few annotated images. On the other hand, transfer learning based methods \cite{LSTD, TFA, MPSR, FSCE} utilize the previous knowledge from the training on the base classes by finetuning the pre-trained model on the novel classes. Nevertheless, most of the aforementioned approaches have ignored the catastrophic forgetting of the base categories and have focused on improving the detection performance of the novel classes. But in  many use cases, it is crucial for perception systems to learn new classes while maintaining their performance on the previously seen ones. For example, in a pick and place task, a robot should be able to learn new objects without forgetting how to handle the old ones.

For this purpose, the task of detecting both base and novel classes, named Generalized Few-Shot Object Detection (G-FSOD), was introduced by \cite{TFA, gfsod}. The two-stage finetuning approach \cite{TFA} (TFA) was among the first to tackle the G-FSOD problem. It jointly finetunes the detector on a balanced set of base and novel classes in a slow learning setting.  In this case, only the classification and box regression layers are finetuned while keeping the backbone and the RPN frozen. Although this has resulted in less forgetting, it limits the performance on novel classes.  As a remedy, Fan \etal \cite{gfsod} has proposed Retentive R-CNN, a transfer-learning-based approach that eliminates the forgetting of the base classes by utilizing an ensemble of base and novel models during inference. Although it succeeds in preventing the performance drop on base classes, the model capacity is increased along with inference time and memory. Furthermore, it restricts the knowledge transfer to the novel task resulting in a lower performance. 

Another line of work that shares a similar interest in alleviating catastrophic forgetting is the continual learning (CL) paradigm.  The key objective of CL methods is to accumulate knowledge across previous tasks to quickly learn new ones without forgetting. Following \cite{cl_survey}, the three main CL approaches are categorized as replay-based, regularization-based, and parameter isolation methods. Replay methods \cite{iCarl, dgr, er, ser, tem, cope, pr, cclugm, lgm, gem, agem, MER, gss} store or generate samples from previous tasks to be replayed while learning a new task. In contrast, regularization-based methods \cite{ewc, lwf, si, lfl, ebll, dmc} merely introduce a regularization term to the objective function to incorporate knowledge from previous tasks without storing any data. Parameter isolation methods \cite{packnet, pnn, piggyback, hat, expert_gate, rcl, dan} assign separate model parameters to each task eliminating forgetting. Nonetheless, CL methods have not yet been exploited in the context of G-FSOD. 

Inspired by the CL replay-based methods, this paper proposes a Constrained finetuning Approach (CFA) to allow better transfer of knowledge between base and novel tasks. CFA extends a replay-based constrained optimization CL method, namely Average Gradient Episodic Memory (A-GEM) \cite{agem}, to G-FSOD. In A-GEM, catastrophic forgetting is considered to occur when the angle between the loss gradient vectors of previous tasks and the proposed gradient update for the current tasks is obtuse. Hence,  A-GEM projects the proposed gradients for the current task orthogonal to the previous task gradients in case of violation. However, starting from the assumption that this may hinder effective knowledge exchange between tasks in G-FSOD, we further constrain the objective function deriving a new gradient update rule. Extensive experiments on MS-COCO and PASCAL-VOC show, that our CFA achieves state-of-the-art performance in the G-FSOD and FSOD settings (as shown in \cref{fig:teaser}).

\section{Related Works}
\label{sec:related_works}
\textbf{Object Detection.} There are two main types of object detectors: two-stage and one-stage detectors. The former \cite{R-CNN, FastR-CNN, FasterR-CNN} feature a proposal generation stage. In Faster R-CNN \cite{FasterR-CNN}, this stage consists of a Region Proposal Network (RPN), which comprises a three-layer CNN that classifies and refines the predicted proposals. Next, the proposals are fed into the classification and localization heads to finally output the detected objects. On the other hand, one-stage detectors \cite{SSD, YOLOv1, YOLOv2, EfficientDet}  directly classify and localize the objects.

\textbf{Few-shot Object Detection.} FSOD approaches can be categorized into two groups: transfer learning-based and meta-learning-based methods. Firstly, the transfer-learning-based frameworks \cite{LSTD, TFA, MPSR} learn to transfer knowledge from source categories to novel categories via finetuning. LSTD \cite{LSTD}, introduces two regularization modules for background suppression and constrained knowledge transfer for less confusion on base classes. TFA \cite{TFA} finetunes only the box predictor on a small balanced training set containing both base and novel classes along with a cosine similarity based box classifier. MPSR \cite{MPSR} tackles the high scale variations by generating multi-scale positive samples as object pyramids to refine the prediction at multiple scales. On the other hand,  meta-learning-based methods \cite{FSRW, MetaDet, MetaRCNN, FSOD-RPN, FsDetView, CME} learns-to-learn solving a set of unrelated tasks. The aim is to perform an exemplar search in the instance level using only a few annotated images support sets. 

\textbf{Generalized Few-shot Object Detection.} G-FSOD \cite{TFA, ONCE, gfsod} is an emerging sub-discipline of FSOD focusing on detecting both the base and novel classes. While TFA \cite{TFA} attempts to avoid forgetting by finetuning on base and novel classes, ONCE \cite{ONCE} tackles the problem in an incremental setting using a meta-learning approach with a CenterNet \cite{center-net} detector. The key idea is to meta-learn a class code generator that incrementally learns to synthesize a class code to the novel classes. Retentive R-CNN \cite{gfsod} proposes a transfer-learning-based approach that eliminates forgetting on base classes. It leverages the base-trained detector head along with the finetuned detector head to detect both classes. Similarly, it integrates both the pre-trained RPN on base classes and another RPN to finetune.

\textbf{Continual Learning.} Recently, continual learning (CL) paradigms have gained more recognition in various computer vision tasks \cite{cl_survey}. Firstly, parameter isolation methods \cite{packnet, piggyback, hat, expert_gate, rcl, dan} tackle the task-incremental setting, requiring the task identity during inference. Secondly, regularization-based methods \cite{ewc, lwf, si, lfl, ebll, dmc} consolidate prior knowledge while learning the new task by regularizing the model parameters. Finally, replay methods \cite{iCarl, dgr, er, ser, tem, cope, pr, cclugm, lgm, gem, agem, MER, gss} leverages stored samples from previous tasks \cite{iCarl} or generates pseudo-samples using a generative model \cite{dgr}. The forgetting is mitigated by replaying the samples in a joint training form with the new task samples. Moreover, constrained methods \cite{gem, agem, MER, gss} alleviate the forgetting of previous tasks by constraining the optimization of the new task loss. Specifically, gradient episodic memory (GEM) \cite{gem} and average gradient episodic memory (A-GEM) \cite{agem} avoid the joint training between the new and old tasks by guiding the new task's optimization process. Recently, Riemer \etal \cite{MER} has proposed an experience replay algorithm that utilizes a meta-learning optimization approach. In \cite{gss}, a selection scheme for storing necessary samples from previous tasks is proposed.

\section{Approach}
In this section, the G-FSOD problem is formally introduced. Next, we revisit a continual learning approach, namely A-GEM\cite{agem}, and show how it can be integrated into the context of G-FSOD. Finally, we present our CFA algorithm which extends A-GEM by further constraining the gradient update on the novel task to foster a mutual knowledge transfer across tasks.   

\subsection{Problem Formulation}
In FSOD and G-FSOD, a dataset is split into a base dataset $\mathcal{D}_b$ containing abundant examples of base classes $\mathcal{C}_b$, and a novel dataset $\mathcal{D}_n$ comprising a handful examples of novel classes $\mathcal{C}_n$ (i.e., $\mathcal{C}_b \cap \mathcal{C}_n = \phi$). Formally, $\mathcal{D}_b=\{\ (x, y)\ |~y = \{(c_{i},b_{i})\}, c_{i} \in \mathcal{C}_{b}\}$, $\mathcal{D}_n=\{\ (x, y)\ |~y = \{(c_{i},b_{i})\}, c_{i} \in \mathcal{C}_{n}\}$, where $x \in \mathcal{X}$ is an input image, and  $y \in \mathcal{Y}$ is the corresponding annotation. $c_i$ and $b_i$ are  the class label and bounding box coordinates of each instance $i$ in the image, respectively. The objective of G-FSOD task is to find a hypothesis $h(\cdot)$ in the hypothesis space that is able to learn the novel classes without forgetting the previously seen base classes.

The few-shot training comprises of two stages: base and novel training. The former trains on $\mathcal{C}_b$ with abundant examples whereas the latter leverages only a handful of examples from $\mathcal{C}_n$. The novel training scheme varies depending on the task. In FSOD, only a few-shots of $\mathcal{C}_n$ are utilized, while in G-FSOD a balanced set from both categories is used. 
Although in G-FSOD, the model has the advantage of interacting with the base classes to avoid any performance drop, we argue that the few-shots can not perfectly represent the base data distribution, leading to overfitting on the limited examples. This has been shown in previous works \cite{TFA, gfsod, defrcn}, where better performance is achieved on the base task with degradation on the novel task. In this work, we seek to bridge the performance gap between G-FSOD and FSOD. More specifically, we strive to alleviate the catastrophic forgetting without increasing the model capacity or hindering novel classes' performance.

As previously mentioned in \cref{sec:related_works}, there are two main approaches of FSOD: finetuning based and meta-learning based. While finetuning approaches can achieve lower performance on $\mathcal{C}_n$ compared to meta-based detectors, they tend to perform better on $\mathcal{C}_b$. Moreover, they offer more flexibility allowing for easier integration into different detection frameworks. One finetuning method, namely Retentive-R-CNN \cite{gfsod}, has eliminated forgetting by reusing a frozen RPN and RoI head from the base task; however, the performance declined on the novel task. To remedy this issue, we present an alternative finetuning approach for G-FSOD.

\subsection{Revisiting the Gradient Episodic Memory Algorithm}
\label{subsec:constrained_approach}
Inspired by the replay-based continual learning (CL) approaches \cite{gem, agem}, the proposed CFA alters the search strategy to find optimal parameters that lead to a better generalization across tasks. This approach has a two-fold advantage: it does not need any tailored data augmentations or specific model modifications while being flexible enough to integrate into various detectors independent of their architecture. Specifically, CFA regularizes the gradient update via a subset of the base dataset stored in the episodic memory, analogous to A-GEM \cite{agem}. 

Episodic memory based approaches \cite{gem, agem, MER, gss} aim to alleviate catastrophic forgetting by maintaining an episodic memory $\mathcal{M}_k$ for each task $k$. Not only do they prevent the losses of the previous tasks from increasing, but they also allow their decrease resulting in a positive backward transfer. Meaning the performance on previous tasks can be improved while learning new ones. Rather than optimizing for all the samples in the episodic memory as originally proposed \cite{gem} instead, A-GEM \cite{agem} tries to guarantee that the average episodic memory loss does not increase over a mini-batch of samples from the episodic memory. 

Firstly, we show how A-GEM \cite{agem} can be exploited in the context of G-FSOD. The employed episodic memory $\mathcal{M}_{b}$ stores few-shots $K$ of the base classes that are randomly sampled from $\mathcal{D}_b$, where $K$ is chosen to match the number few-shots in the novel set $\mathcal{D}_n$. In contrast to the CL scheme, we train batch-wise where the data is seen more than once and is not handled one by one on-the-fly. Moreover, during novel training, the episodic memory is static meaning that no further samples are added. 

The finetuning is then conducted as follows: a mini-batch is sampled from $\mathcal{M}_{b}$ to compute the base gradient, $g_b$. Then, another mini-batch is sampled from $\mathcal{D}_n$ and the novel gradient, $g_n$, is calculated. As defined in \cite{gem, agem}, a positive knowledge transfer occurs when the angle between $g_b$ and $g_n$ is acute. If this constraint is satisfied, $g_n$ is back-propagated as it is. Otherwise, $g_n$ is projected to a region in the hypothesis space closer to the base task gradients, dictated by $g_b$, before back-propagating on the model. 

We formulate the objective function for the G-FSOD task as follows:  
\begin{align}
\text{minimize}_\theta\quad& \mathcal{L}(h_\theta(x), y)\nonumber\\
\text{subject to}\quad& \mathcal{L}(h_\theta,  \mathcal{M}_{b}) \leq
\mathcal{L}(h^{b}_{\theta},  \mathcal{M}_{b}),
\label{eq:projgrad-batch}
\end{align}
where $(x, y) \in \mathcal{D}_n$. $h^{b}_{\theta}$ is the pretrained model on $\mathcal{D}_b$. The loss function on $\mathcal{M}_{b}$ is given by:

\begin{equation}
\label{eq:previous_task_loss}
    \mathcal{L}(h_{\theta}, \mathcal{M}_b) = \frac{1}{\left| \mathcal{M}_b \right|} \sum_{(x_i,y_i)\in \mathcal{M}_b} \mathcal{L}(h_{\theta}(x_i), y_i).
\end{equation}

 $\mathcal{L}$ is the overall detection loss function and is defined in \cite{FasterR-CNN} as:

\begin{equation}
\label{eq:det_loss}
    \mathcal{L} = \mathcal{L}_{rpn} + \mathcal{L}_{cls} + \mathcal{L}_{reg}, 
\end{equation}
where $\mathcal{L}_{rpn}$ is the RPN loss function.  $\mathcal{L}_{cls}$ and  $\mathcal{L}_{reg}$ are the classification and bounding box regression loss. 

 \begin{figure}[t!]
 \centering
 \includegraphics[width=1.0\linewidth, height=0.48\linewidth]{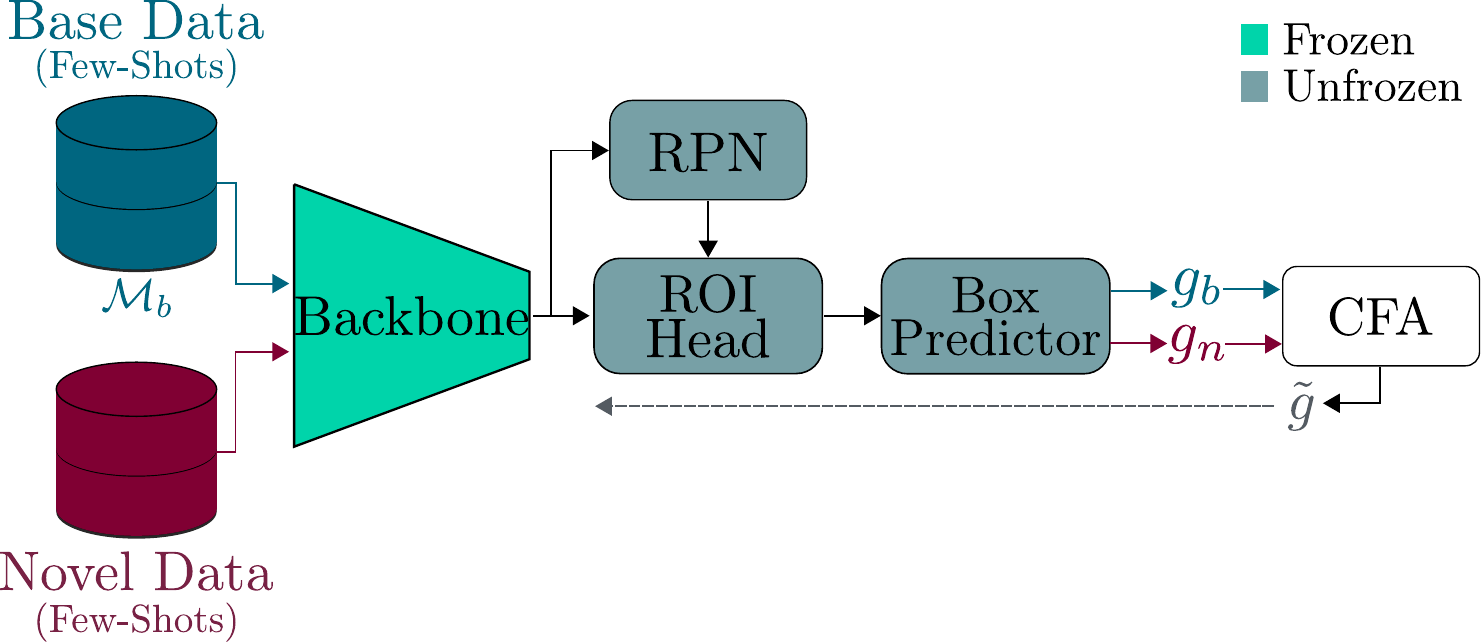}
 \caption{Overview of the fine-tuning stage with CFA on Faster R-CNN \cite{FasterR-CNN}. Gradients are computed from each mini-batch. CFA calculates the final gradient update rule which back-propagates on the unfrozen components.} %
 \vspace{-2mm}
 \label{fig:cfa_gfsod}%
\end{figure}
Analogous to \cite{gem, agem}, the optimization problem in \cref{eq:projgrad-batch} can be reduced to a quadratic programming (QP) problem in the context of A-GEM as follows:
\begin{align}
	 	\textrm{minimize}_{\tilde{g}_n} \quad &\frac{1}{2}||g_n-\tilde{g}_n||_2^2 \nonumber\\
	 	\textrm{subject to} 				\quad &\tilde{g}_n^\top g_{b} \geq 0, \label{eq:qp_1}  
\end{align}
where $\tilde{g}_n$ is the projected gradient update for the novel task. Formally, the closed-form A-GEM gradient update rule is realized as:
\begin{equation}
    \tilde{g}_n =   g_n - \frac{g_{n}^\top g_{b}}{g_{b}^\top g_{b}} \cdot g_{b} ,
\end{equation}
which implies that only in case of violation of the A-GEM constraint, the novel gradient is projected orthogonal to the base gradient. 

\begin{figure*}
 \centering
 \begin{subfigure}{0.3\linewidth}
    \includegraphics[width=0.95\linewidth]{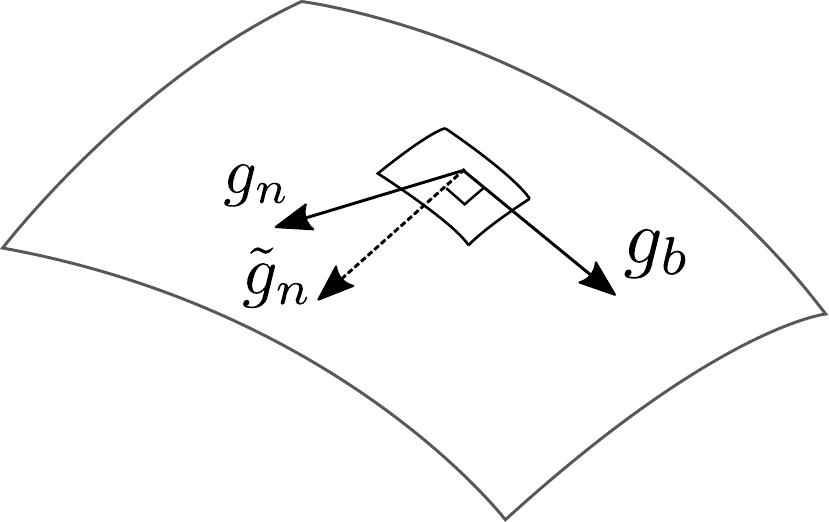}
    \caption{Vanilla A-GEM approach.}
    \label{fig:agem}
 \end{subfigure}
 \begin{subfigure}{0.3\linewidth}
    \includegraphics[width=0.95\linewidth]{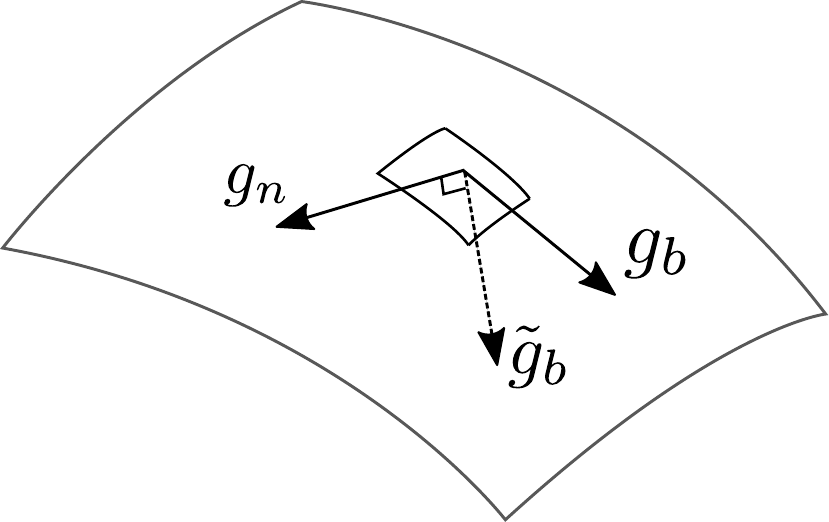}
    \caption{Intermediate base gradient update.}
    \label{fig:intermediate}
 \end{subfigure}
 \begin{subfigure}{0.3\linewidth}
    \includegraphics[width=0.95\linewidth]{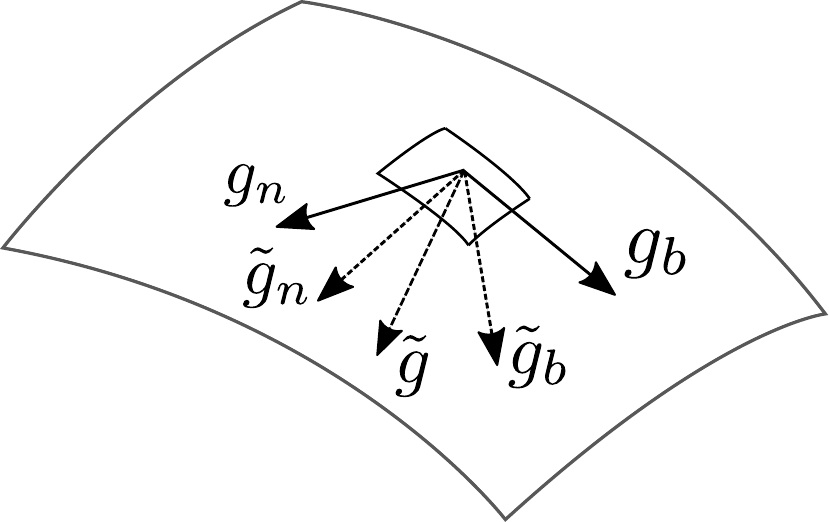}
    \caption{The proposed gradient update via CFA.}
    \label{fig:cfa}
 \end{subfigure}
  \caption{Visualization of the gradient update for vanilla A-GEM and the proposed CFA. In \cref{fig:agem}, we present the novel task gradient update using the A-GEM where the novel task gradient is projected orthogonally to base task gradient. \cref{fig:intermediate} shows the the solution for the proposed constraint where the base task gradient is projected to a right angle with the novel task gradient. Finally, the final gradient update for the CFA algorithm is shown in \cref{fig:cfa}.}%
  \vspace{-1em}
  \label{fig:gradients_vis}%
\end{figure*}

\subsection{Constraint-based Finetuning Approach for G-FSOD}
\label{subsec:cfa}
Compared to finetuning approaches, A-GEM applied in the context G-FSOD, provides a better regularization to the learning process preventing early overfitting. However, we argue that depending merely on the above mentioned constraint might hinder knowledge transfer across tasks, due to the following reasons: (1) the base gradient is only back-propagated in case of violation and thus has little influence during finetuning while learning the novel classes, (2) projecting the novel gradient orthogonally is too restrictive for diverse novel feature learning.
 
Motivated by the aforementioned observations, we propose to minimize the angle between $g_b$ and $g_n$ instead of always projecting $g_n$ orthogonally in case of violation. The proposed CFA algorithm arises from a joint optimization problem on both tasks, where a further constraint is inflicted to account for the performance on base categories. In CFA, both the novel and base gradients are back-propagated. The scheme is illustrated in \cref{fig:cfa_gfsod}. In case of violation, $g_n$ is projected orthogonally to $\tilde{g}_n$ with respect to $g_b$, while $g_b$ is, in turn, projected orthogonally to $\tilde{g}_b$ with respect to $g_n$.
Formally, our proposed constrained optimization problem is formulated as follows:
\begin{align}
	 	\textrm{minimize}_{\tilde{g}_b, \tilde{g}_n} \quad &\frac{1}{2}||g_n-\tilde{g}_n||_2^2 + \frac{1}{2}||g_b-\tilde{g}_b||_2^2 \nonumber\\
	 	\textrm{subject to} 				\quad &\tilde{g}_n^\top g_{b} \geq 0, \nonumber\\ 
	 	                                    \quad &\tilde{g}_b^\top g_{n} \geq 0,   \label{eq:qp_2}  
\end{align}
where $\tilde{g}_b$ and $\tilde{g}_n$ denotes the projected gradient update for the base task and novel task, respectively. Solving the above constrained optimization problem using the method of Lagrange multipliers, the gradient update rules are derived as follows: 
\begin{align}
        \tilde{g}_n =  g_{n} - \Big(\frac{g_{n}^\top g_{b}}{g_{b}^\top g_{b}} \Big) \cdot g_{b},\\
        \tilde{g}_b =  g_{b} - \Big(\frac{g_{b}^\top g_{n}}{g_{n}^\top g_{n}} \Big) \cdot g_{n}.
\end{align}
The formal proof of the aforementioned update rules for CFA is provided in the supplementary material section. 

Instead of performing two gradient updates, a single update rule can be realized by averaging $\tilde{g}_b$ and $\tilde{g}_n$:   
\begin{equation}
\label{eq:update_final}
    \tilde{g} =   \frac{1}{2} \Big(1 -  \frac{g_{n}^\top g_{b}}{g_{b}^\top g_{b}} \Big) \cdot g_{b} +  \frac{1}{2} \Big( 1 - \frac{g_{b}^\top g_{n}}{g_{n}^\top g_{n}} \Big) \cdot g_{n}.
\end{equation}
\cref{eq:update_final} can be interpreted as an adaptive re-weighting of the two gradients $g_b$ and $g_n$ that attempts to balance their contribution during the finetuning stage. The advantage of the CFA algorithm, as shown in \cref{alg:cfa}, is two-fold. Firstly, the base gradients always contribute to the finetuning stage. Secondly, the algorithm searches for the optimal direction of back-propagation by re-weighting the contributions of each gradient while keeping the angle between the last gradient update $\tilde{g}$ and the base gradient $g_b$ less than 90°. A visualization of the differences between A-GEM and CFA gradient update rules is shown in \cref{fig:gradients_vis}.

\setlength{\textfloatsep}{15pt}
\begin{algorithm}[t!]
\caption{CFA} \label{alg:cfa}
\footnotesize
\small
\begin{algorithmic}[1]
  \Procedure{TRAIN}{$f_{\theta}, \mathcal{D}_b, \mathcal{D}_n$}
    \State $\mathcal{M}_b \sim \mathcal{D}_b$
            \For{$n_{epoch} = 1, \ldots, N_{epoch}$}:
                \For{$(x_{n},y_{n})$ in $\mathcal{D}_n$}
                        \State $(x_{b}, y_{b})\sim\mathcal{M}_b$
                        \State $g_{b}\gets\nabla_{\theta} \mathcal{L}(f_{\theta}(x_{b}), y_{b})$
                        \State $g_{n}\gets\nabla_{\theta} \mathcal{L}(f_{\theta}(x_{n}), y_{n})$
                        \If{$g_{n}^\top g_{b} \geq 0$}
                            \State $\tilde{g}\gets \frac{g_{n} + g_{b}}{2}$
                        \Else
                            \State $\tilde{g}\gets \frac{1}{2} \Big(1 -  \frac{g_{n}^\top g_{b}}{g_{b}^\top g_{b}} \Big) \cdot g_{b} +  \frac{1}{2} \Big( 1 - \frac{g_{b}^\top g_{n}}{g_{n}^\top g_{n}} \Big) \cdot g_{n}$
                        \EndIf
                        \State $\theta \gets \theta - \eta\tilde{g}$
                \EndFor
            \EndFor
    \State \textbf{return} $f_{\theta}$
  \EndProcedure
\end{algorithmic} 
\end{algorithm}

\begin{table*}
    \centering
    \begin{tabular}{c | c | c c c | c c c | c c c}
      \Xhline{1pt}
      \multirow{2}{*}{\textbf{Methods} / \textbf{Shots}} & \multirow{2}{*}{\textbf{w/E}} &
      \multicolumn{3}{c |}{\textbf{5 shot}}  &
      \multicolumn{3}{c |}{\textbf{10 shot}}  &
      \multicolumn{3}{c}{\textbf{30 shot}}\\
      & & \textbf{AP} &  \textbf{bAP} & \textbf{nAP} & \textbf{AP} &  \textbf{bAP} & \textbf{nAP} & \textbf{AP} &  \textbf{bAP} & \textbf{nAP} \\ \Xhline{1pt}
      FRCN-ft-full\cite{TFA} & \xmark & 18.0 & 22.0 & 6.0 & 18.1 & 21.0 & 9.2 & 18.6 & 20.6 & 12.5 \\
      TFA w/ fc\cite{TFA} & \xmark & 27.5 & 33.9 & 8.4 & 27.9 & 33.9 & 10.0 & 29.7 & 35.1 & 13.4 \\
      TFA w/ cos\cite{TFA} & \xmark & 28.1 & 34.7 & 8.3 & 28.7 & 35.0 & 10.0 & 30.3 & 35.8 & 13.7 \\
      MPSR\cite{MPSR} & \xmark & - & - & - & 15.3 & 17.1 & 9.7 & 17.1 & 18.1 & 14.1 \\
      DeFRCN\cite{defrcn} & \xmark & 28.7 & 33.1 & \second{15.3} & 30.6 & 34.6 & \second{18.6} & 31.6 & 34.7 & \second{22.5}\\
      ONCE~\cite{ONCE} & \xmark & 13.7 & 17.9 & 1.0 & 13.7 & 17.9 & 1.2 & - & - & -  \\
      Meta R-CNN$^{*}$\cite{MetaRCNN}& \xmark & 3.6 & 3.5 & 3.8 & 5.4 & 5.2 & 6.1 & 7.8 & 7.1 & 9.9 \\
      FSRW\cite{FSRW}& \xmark & - & - & - & - & - & 5.6 & - & - & 9.1 \\
      FsDetView$^{*}$\cite{FsDetView}& \xmark & 5.9 & 5.7 & 6.6 & 6.7 & 6.4 & 7.6 & 10.0 & 9.3 & 12.0 \\
      \Xhline{1pt}
      \rowcolor[HTML]{EFEFEF}
      CFA w/ fc & \xmark & \best{30.1} & \best{37.1} & 9.0 & \second{30.8} & \best{37.6} & 10.5 & \second{31.9} & \best{37.7} & 14.7\\
      \rowcolor[HTML]{EFEFEF}
      CFA w/ cos& \xmark & \second{29.7} & \second{36.3} & 9.8 & 30.3 & \second{36.6} & 11.3 & 31.7 & \second{37.0} & 15.6 \\
      \rowcolor[HTML]{EFEFEF}
      CFA-DeFRCN& \xmark & \best{30.1} & 35.0 & \best{15.6} & \best{31.4} & 35.5 & \best{19.1} & \best{32.0} & 35.0 & \best{23.0} \\
      \midrule[1.5pt]
      Retentive R-CNN\cite{gfsod}& \cmark & 31.5 & \second{39.2} & 8.3 & 32.1 & 39.2 & 10.5 & 32.9 & \second{39.3} & 13.8 \\
      \Xhline{1pt}
      \rowcolor[HTML]{EFEFEF}
      CFA w/ fc & \cmark & 31.8 & \best{39.5} & 8.8 & 32.2 & \best{39.5} & 10.4 & 33.2 & \best{39.5} & 14.3\\
      \rowcolor[HTML]{EFEFEF}
      CFA w/ cos& \cmark & \second{32.0} & \best{39.5} & \second{9.6} & \second{32.4} & \second{39.4} & \second{11.3} & \second{33.4} & \best{39.5} & \second{15.1} \\
      \rowcolor[HTML]{EFEFEF}
      CFA-DeFRCN& \cmark & \best{33.0} & 38.9 & \best{15.6} & \best{34.0} & 39.0 & \best{18.9} & \best{34.9} & 39.0 & \best{22.6} \\
      \Xhline{1pt}
    \end{tabular}
    \caption{G-FSOD experimental results for 5,10,30-shot settings on MS-COCO. We report AP, bAP, nAP for all, base, and novel classes, respectively. w/E denotes whether the ensemble-learning based evaluation protocol of Retentive-RCNN \cite{gfsod} was used. Colored results represent the {\color{red}best} and {\color{blue}second-best}. '*' represents results reported in \cite{gfsod} and \cite{defrcn}. '-' denotes unreported results in the original work.}
    \label{tab:coco}
\end{table*}
\begin{table*}
\footnotesize
    \centering
    \begin{tabular}{c| c | c c c c c | c c c c c | c c c c c}
      \Xhline{1pt}
      \multirow{2}{*}{\textbf{Methods} / \textbf{Shots}} &
      \multirow{2}{*}{\textbf{w/E}} &
      \multicolumn{5}{c |}{\textbf{All Set 1}}  &
      \multicolumn{5}{c |}{\textbf{All Set 2}}  &
      \multicolumn{5}{c}{\textbf{All Set 3}}\\

      & & 1 &  {2} & {3} & {5} &  {10} & 1 &  {2} & {3} & {5} &  {10} & 1 &  {2} & {3} & {5} &  {10} \\ \Xhline{1pt}
      FRCN-ft-full\cite{TFA} &\xmark & 55.4 & 57.1 & 56.8 & 60.1 & 60.9 & 50.1 & 53.7 & 53.6 & 55.9 & 55.5 & 58.5 & 59.1 & 58.7 & 61.8 & 60.8 \\
      TFA w/ fc\cite{TFA}&\xmark & 69.3 & 66.9 & 70.3 & 73.4 & 73.2 & 64.7 & 66.3 & 67.7 & 68.3 & 68.7 & 67.8 & 68.9 & 70.8 & 72.3 & 72.2 \\
      TFA w/ cos\cite{TFA}&\xmark & 69.7 & 68.2 & 70.5 & 73.4 & 72.8 & 65.5 & 65.0 & 67.7 & 68.0 & 68.6 & 67.9 & 68.6 & 71.0 & 72.5 & 72.4 \\
      MPSR\cite{MPSR}&\xmark & 56.8 & 60.4 & 62.8 & 66.1 & 69.0 &
       53.1 & 57.6 & 62.8 & 64.2 & 66.3 & 55.2 & 59.8 & 62.7 & 66.9 & 67.7 \\
      DeFRCN\cite{defrcn}&\xmark & \second{73.1} & \second{73.2} & \second{73.7} & \second{75.1} & \best{74.4} &
      \second{68.6} & \second{69.8} & \second{71.0} & \second{72.5} & \second{71.5} & \second{72.5} & \second{73.5} & \second{72.7} & \best{74.1} & \second{73.9} \\
       Meta R-CNN$^{*}$\cite{MetaRCNN}&\xmark & 17.5 & 30.5 & 36.2 & 49.3 & 55.6 & 19.4 & 33.2 & 34.8 & 44.4 & 53.9 & 20.3 & 31.0 & 41.2 & 48.0 & 55.1 \\
       FSRW\cite{FSRW}&\xmark & 53.5 & 50.2 & 55.3 & 56.0 & 59.5 & 55.1 & 54.2 & 55.2 & 57.5 & 58.9 & 54.2 & 53.5 & 54.7 & 58.6 & 57.6 \\
       FsDetView$^*$\cite{FsDetView}&\xmark & 36.4 & 40.3 & 40.1 & 50.0 & 55.3 & 36.3 & 43.7 & 41.6 & 45.8 & 54.1 & 37.0 & 39.5 & 40.7 & 50.7 & 54.8 \\ \Xhline{1pt}
       \rowcolor[HTML]{EFEFEF}
        CFA w/ fc&\xmark & 69.5 & 68.2 & 69.8 & 73.5 & \second{74.3} & 66.0 & 66.9 & 69.2 & 70.1 & 71.1 & 67.7 & 69.0 & 70.9 & 72.6 & 73.5 \\
        \rowcolor[HTML]{EFEFEF}
        CFA w/ cos&\xmark & 69.1 & 69.8 & 71.9 & 73.6 & 73.9 & 64.8 & 66.5 & 68.3 & 69.5 & 70.5 & 67.7 & 69.7 & 71.9 & \second{73.0} & 73.5 \\
       \rowcolor[HTML]{EFEFEF}
        CFA-DeFRCN&\xmark & \best{73.8} & \best{74.6} & \best{74.5} & \best{76.0} & \best{74.4} & \best{69.3} & \best{71.4} & \best{72.0} & \best{73.3} & \best{72.0} & \best{72.9} & \best{73.9} & \best{73.0} & \best{74.1} & \best{74.6} \\ \midrule[1.5pt]
       Retentive R-CNN\cite{gfsod}&\cmark & 71.3 & \second{72.3} & 72.1 & 74.0 & 74.6 & 66.8 & \second{68.4} & 70.2 & 70.7 & 71.5 & 69.0 & 70.9 & 72.3 & 73.9 & 74.1 \\
       \Xhline{1pt}
        \rowcolor[HTML]{EFEFEF}
       CFA w/ fc&\cmark & 70.3 & 69.5 & 71.0 & 74.4 & 74.9 & \second{67.0} & 68.0 & 70.2 & 70.8 & 71.5 &  69.1 & 70.1 & 71.6 & 73.3 & \second{74.7} \\
       \rowcolor[HTML]{EFEFEF}
       CFA w/ cos&\cmark & \second{71.4} & 71.8 & \second{73.3} & \second{74.9} & \second{75.0} & 66.8 & \second{68.4} & \second{70.4} & \second{71.1} & \second{71.9} & \second{69.7} & \second{71.2} & \second{72.6} & \second{74.0} & \second{74.7} \\
       \rowcolor[HTML]{EFEFEF}
       CFA-DeFRCN&\cmark & \best{75.0} & \best{76.0} & \best{76.8} & \best{77.3} & \best{77.3} & \best{70.4} & \best{72.7} & \best{73.7} & \best{74.7} & \best{74.2} & \best{74.7} & \best{75.5} & \best{75.0} & \best{76.2} &  \best{76.6} \\ 
        \Xhline{1pt}
    \end{tabular}
    \caption{G-FSOD experimental results for 1,2,3,5,10-shot settings on the three all sets of Pascal VOC (AP50). w/E denotes whether the ensemble-learning based evaluation protocol of Retentive-RCNN \cite{gfsod} was used. Colored results represent the {\color{red}best} and {\color{blue}second-best}. '*' represents results reported in \cite{gfsod} and \cite{defrcn}.}
    \label{tab:voc-all}
    \vspace{-2em}
\end{table*}
\begin{table*}
\footnotesize
    \centering
    \begin{tabular}{c| c | c c c c c | c c c c c | c c c c c}
      \Xhline{1pt}
      \multirow{2}{*}{\textbf{Methods} / \textbf{Shots}} &
      \multirow{2}{*}{\textbf{w/E}} &
      \multicolumn{5}{c |}{\textbf{Novel Set 1}}  &
      \multicolumn{5}{c |}{\textbf{Novel Set 2}}  &
      \multicolumn{5}{c}{\textbf{Novel Set 3}}\\
      & & 1 &  {2} & {3} & {5} &  {10} & 1 &  {2} & {3} & {5} &  {10} & 1 &  {2} & {3} & {5} &  {10} \\ \Xhline{1pt}
      FRCN-ft-full\cite{TFA} & \xmark & 15.2 & 20.3 & 29.0 & 25.5 & 28.7 & 13.4 & 20.6 & 28.6 & 32.4 & 38.8 & 19.6 & 20.8 & 28.7 & 42.2 & 42.1 \\
      TFA w/ fc\cite{TFA}& \xmark & 36.8 & 29.1 & 43.6 & 55.7 & 57.0 & 18.2 & 29.0 & 33.4 & 35.5 & 39.0 & 27.7 & 33.6 & 42.5 & 48.7 & 50.2 \\
      TFA w/ cos\cite{TFA}& \xmark & 39.8 & 36.1 & 44.7 & 55.7 & 56.0 & 23.5 & 26.9 & 34.1 & 35.1 & 39.1 & 30.8 & 34.8 & 42.8 & 49.5 & 49.8 \\
      MPSR\cite{MPSR}& \xmark & 42.8 & 43.6 & 48.4 & 55.3 & 61.2 & 29.8 & 28.1 & 41.6 & 43.2 & 47.0 & 35.9 & 40.0 & 43.7 & 48.9 & 51.3 \\
      DeFRCN\cite{defrcn}& \xmark & \second{57.0} & \second{58.6} & \second{64.3} & \second{67.8} & \second{67.0} &
     \second{35.8} & \second{42.7} & \second{51.0} & \second{54.4} & \second{52.9} & \second{52.5} & \second{56.6} & \second{55.8} & \best{60.7} & \second{62.5} \\ \Xhline{1pt}
      Meta R-CNN$^{*}$\cite{MetaRCNN}& \xmark & 16.8 & 20.1 & 20.3 & 38.2 & 43.7 & 7.7 & 12.0 & 14.9 & 21.9 & 31.1 & 9.2 & 13.9 & 26.2 & 29.2 & 36.2 \\
      FSRW\cite{FSRW}& \xmark & 14.8 & 15.5 & 26.7 & 33.9 & 47.2 & 15.7 & 15.3 & 22.7 & 30.1 & 39.2 & 19.2 & 21.7 & 25.7 & 40.6 & 41.3 \\
      MetaDet\cite{MetaDet}& \xmark & 18.9 & 20.6 & 30.2 & 36.8 & 49.6 & 21.8 & 23.1 & 27.8 & 31.7 & 43.0 & 20.6 & 23.9 & 29.4 & 43.9 & 44.1\\
      FsDetView$^*$\cite{FsDetView}& \xmark & 25.4 & 20.4 & 37.4 & 36.1 & 42.3 & 22.9 & 21.7 & 22.6 & 25.6 & 29.2 & 32.4 & 19.0 & 29.8 & 33.2 & 39.8 \\ \Xhline{1pt}
      \rowcolor[HTML]{EFEFEF}
      CFA w/ fc& \xmark & 40.0 & 35.5 & 40.9 & 54.1 & 56.9 & 22.2 & 27.1 & 35.2 & 38.5 & 40.9 & 29.7 & 35.1 & 39.5 & 47.2 & 51.3 \\
      \rowcolor[HTML]{EFEFEF}
      CFA w/ cos& \xmark & 41.2 & 43.6 & 49.5 & 56.5 & 57.3 & 21.3 & 27.4 & 35.3 & 39.1 & 42.1 & 31.7 & 39.1 & 44.6 & 49.9 & 52.6 \\
      \rowcolor[HTML]{EFEFEF}
    CFA-DeFRCN& \xmark & \best{58.2} & \best{63.3} & \best{65.8} & \best{68.9} & \best{67.1} & \best{37.1} & \best{45.5} & \best{51.3} & \best{55.2} & \best{53.8} & \best{54.7} & \best{57.8} & \best{56.9} & \second{60.0} & \best{63.3} \\
      \midrule[1.5pt]
      Retentive R-CNN\cite{gfsod}& \cmark& \second{42.4} & \second{45.8} & 45.9 & 53.7 & 56.1 & 21.7 & \second{27.8} & 35.2 & 37.0 & 40.3 & 30.2 & 37.6 & 43.0 & 49.7 & 50.1 \\
      \Xhline{1pt}
      \rowcolor[HTML]{EFEFEF}
      CFA w/ fc& \cmark & 39.0 & 34.9 & 41.4 & 54.8 & 57.0 & \second{21.8} & 26.1 & \second{35.3} & 37.1 & 40.1 & 29.9 & 34.3 & 40.1 & 47.0 & 52.6 \\
      \rowcolor[HTML]{EFEFEF}
      CFA w/ cos& \cmark &  \second{42.4} & 43.9 & \second{50.3} & \second{56.6} & \second{57.3} & 21.0 & 27.5 & \second{35.3} & \second{38.6} & \second{41.4} &  \second{32.3} & \second{38.0} & \second{44.5} & \second{49.8} & \second{52.7} \\
      \rowcolor[HTML]{EFEFEF}
      CFA-DeFRCN& \cmark & \best{59.0} & \best{63.5} & \best{66.4} & \best{68.4} & \best{68.3} & \best{37.0} & \best{45.8} & \best{50.0} & \best{54.2} & \best{52.5} & \best{54.8} & \best{58.5} & \best{56.5} & \best{61.3} & \best{63.5} \\
      \Xhline{1pt}
    \end{tabular}
    \caption{G-FSOD experimental results for 1,2,3,5,10-shot settings on the three novel sets of Pascal VOC (nAP50). Colored results represent the {\color{red}best} and {\color{blue}second-best}. '*' represents results reported in \cite{gfsod} and \cite{defrcn}. We achieve state-of-the-art novel performance (nAP50) across the three different splits under different few-shot settings.}
    \vspace{-2em}
    \label{tab:voc-novel}
\end{table*}

\section{Experiments}
To evaluate our approach, we follow the well-established G-FSOD benchmarks \cite{TFA, gfsod, defrcn}, where experiments on PASCAL VOC \cite{pascalvoc} and MS-COCO \cite{coco} datasets are conducted. For a fair comparison, we use the same class and data splits as in previous works \cite{FSRW, TFA, MPSR}. 

\subsection{Experimental setting}

\textbf{PASCAL VOC.} The dataset features three different sets, where each comprises 20 categories. Moreover, the classes are split randomly into 15 and 5, base and novel classes, respectively. The data is sampled from both the VOC 2007 and VOC 2012 train/val set for base and novel training. For testing, the VOC 2007 test set is utilized. The results are reported for $K= 1,2,3,5,10,30$ shots.

\textbf{MS-COCO.} The dataset consists of 80 classes where we use 60 base categories disjoint with VOC, while the remaining 20 are used as novel classes. During testing, we use the $5k$ images from the validation set, while the rest is utilized for training. We report the results for $K=5, 10, 30$ shots. 

\textbf{Evaluation setting.}  We consider the G-FSOD evaluation protocol, where a few-shots of the base classes are utilized during novel training. The main aim of G-FSOD is to assess the overall performance of the base and novel categories, which is the more practical case for computer vision tasks. For evaluation metrics, we report AP, bAP, nAP denoting the overall, base, and novel mean average precision, respectively. For PASCAL VOC, only the AP50, bAP50, and nAP50 are utilized.

\textbf{Implementation details.} We adopt Faster R-CNN \cite{FasterR-CNN} as a primary detection framework using a ResNet-101 \cite{resnets} as a backbone and a Feature Pyramid Network (FPN) \cite{fpn}. The learning rate is set to $0.02$ for base training and $0.001$ for novel training.  We optimize the model with a stochastic gradient descent (SGD) optimizer using a momentum of $0.9$ and a weight decay of $0.0001$. The batch size is set to $16$ across all experiments using $4$ Nvidia GeForce 1080Ti GPUs. 

Analogous to TFA \cite{TFA}, we evaluate our method with a fully-connected base classifier (CFA w/fc) and a cosine similarity-based box classifier (CFA w/cos). Moreover, we apply CFA to the state-of-the-art finetuning-based approach DeFRCN \cite{defrcn}, referred to as CFA-DeFRCN, and we follow the original  hyperparameters of the paper. Unlike CFA w/fc and CFA w/cos, CFA-DeFRCN does not employ an FPN, analogous to the baseline DeFRCN. 

Retentive R-CNN \cite{gfsod} has introduced a model-growth based evaluation protocol where the base RPN (RPN$_{b}$) and base detector (DET$_{b}$) are leveraged during inference along with the finetuned novel RPN (RPN$_{n}$) and the novel detector (DET$_{n}$). Specifically, proposals are generated from RPN$_{b}$ and RPN$_{n}$ based on maximum objectness score, concurrently. Next, the proposals are fed to DET$_{b}$ and DET$_{n}$ where detections from DET$_{b}$ (with added $0.1$ bonus) are enouraged for $\mathcal{C}_b$ via a non-maximum suppression. For fair comparison, we further evaluate our methods (CFA w/fc, CFA w/cos, and CFA-DeFRCN) using the aforementioned evaluation protocol, similar to Retentive-RCNN \cite{gfsod}. Both evaluation protocols are illustrated in the supplementary materials.

\subsection{Comparison Experiments}

We perform quantitative comparisons against both transfer-learning \cite{TFA, MPSR, gfsod, defrcn} and meta-learning \cite{ONCE, MetaRCNN, FSRW, FsDetView} based methods under G-FSOD setting. We also compare to Retentive-RCNN using their evaluation protocol \cite{gfsod}. To show the effectiveness of our method, we report our results for CFA w/fc, CFA w/cos, and CFA-DeFRCN.  

\textbf{Results on MS-COCO.} We also evaluate our approach on MS-COCO in \cref{tab:coco} across $K=5,10,30$-shot settings. The standard MS-COCO mean average precision (mAP) metric is used for the base (bAP) and novel (nAP) categories. In the G-FSOD setting, the results show that, independent of the architecture, CFA can help achieve less forgetting on the base categories compared to TFA, while improving on the novel tasks. \cref{tab:coco} shows that CFA-DeFRCN consistently achieves the best overall and novel performance across the three few-shot configurations, compared to both FSOD and G-FSOD methods. 

Furthermore, we report the results of our method using the Retentive-RCNN evaluation protocol. In comparison to Retentive-RCNN \cite{gfsod}, the CFA-finetuned models achieve better performance on both base and novel classes. CFA w/cos performs slightly better than CFA w/fc. This finding is in line with previous observations in TFA \cite{TFA} and Retentive-RCNN\cite{gfsod}, where cosine similarity classifiers allow better generalization due to their robustness against variations of feature norms between base and novel classes. 

Although the ensemble-based evaluation protocol leads to less forgetting on the base classes, it suffers from increased inference time and model capacity (additional experiments are presented in the supplementary material). As shown in \cref{tab:coco}, single CFA-finetuned methods can achieve very comparable performance to Retentive-RCNN \cite{gfsod} on base classes and superior results on novel classes. In this ensemble setting, CFA-DeFRCN does not only employ base and novel RPN and detectors but also base and novel backbones. This is due to the fact that the gradient decoupling layer allows finetuning the backbone with gradients back-propagated from the RoI head.

\textbf{Results on PASCAL VOC.} The overall and novel performance of the G-FSOD models on PASCAL VOC dataset are shown in \cref{tab:voc-all} and \cref{tab:voc-novel}, respectively. The effectiveness of CFA is highlighted across different splits, where single CFA-finetuned models are able to generalize better than the ensemble of base and novel models used in Retentive R-CNN \cite{gfsod}. Additional results on MS-COCO and PASCAL VOC can be found in the supplementary material.  

\subsection{Ablation Study}
To further assess the capability of our proposed CFA algorithm, ablation experiments are conducted on the MS-COCO dataset with a $K=10$-shot scenario.
\begin{table}[t!] \centering
	\setlength{\tabcolsep}{2.5mm}
	\scalebox{0.8}
	{\begin{tabular}{c|ccc|ccc}
			\toprule[1.1pt]
			\multicolumn{1}{c|}{}            & \multicolumn{3}{c|}{}                  &\multicolumn{3}{c}{} \\
			\multicolumn{1}{c|}{\multirow{-2}{*}{Method}}      &\multicolumn{1}{c}{\multirow{-2}{*}{Backone}} & \multicolumn{1}{c}{\multirow{-2}{*}{RPN}}        & \multicolumn{1}{c|}{\multirow{-2}{*}{RoI Head}}                       & \multicolumn{1}{c}{\multirow{-2}{*}{AP}}            & \multicolumn{1}{c}{\multirow{-2}{*}{bAP}}  &  \multicolumn{1}{c}{\multirow{-2}{*}{nAP}}                 \\
			\midrule[0.9pt]
			  &&&& 27.9 & 33.9 & 10.0 \\
			  & &\cmark&&29.9&37.2& 7.9\\
			  &  &&\cmark&28.9&35.4&9.6 \\
			  &&\cmark&\cmark&28.9&35.1&10.2 \\
			\multirow{-5}{*}{TFA}  &\cmark &\cmark&\cmark&24.1&29.0&9.2 \\
			\midrule[0.9pt]
			  & &&&29.6&36.0&10.4\\
			  & &\cmark&&30.3&37.4&9.3 \\
			  &  &&\cmark&\textbf{30.8}&\textbf{37.8}&9.6 \\
			  & &\cmark&\cmark&\textbf{30.8} & 37.6 & \textbf{10.5} \\
			\multirow{-5}{*}{CFA}  &\cmark &\cmark&\cmark&23.9&28.6&10.1 \\

			\bottomrule[1.1pt]
	\end{tabular}}
	\caption{Effect of unfreezing different components while finetuning with CFA in comparison to TFA \cite{TFA}. \cmark\ denotes unfreezing a component. We adopt the FC classifier based detector model \cite{FasterR-CNN}. The results are reported for MS-COCO under 10-shots.}
	\vspace{-1em}
	\label{tab:ablation-study}
\end{table}

\textbf{Impact of unfreezing different components.} We study the influence of various model components in the G-FSOD setting. The results are presented in \cref{tab:ablation-study}.
Firstly, we notice that unfreezing either the RPN or the RoI Head alone leads to suboptimal results. Although the bAP tends to slightly increase compared to the frozen model, there exists a slight drop in the nAP, because the unfrozen component overfits the few novel shots. Secondly, we get the best results in both TFA and CFA by unfreezing the RPN and RoI head. Unfreezing the backbone however leads to deteriorated results. Thirdly, CFA is able to better guide the gradients in the increased search space when unfreezing the RPN and RoI head.  

\begin{table}[t!] \centering
	\setlength{\tabcolsep}{4.0mm}
	\scalebox{0.8}
	{\begin{tabular}{c|c|ccc}
			\toprule[1.1pt]
			\multicolumn{1}{c|}{}            & \multicolumn{1}{c|}{}                  &\multicolumn{3}{c}{} \\
			\multicolumn{1}{c|}{\multirow{-2}{*}{Method}}      &\multicolumn{1}{c|}{\multirow{-2}{*}{Base-Shots}}                        & \multicolumn{1}{c}{\multirow{-2}{*}{AP}}            & \multicolumn{1}{c}{\multirow{-2}{*}{bAP}}  &  \multicolumn{1}{c}{\multirow{-2}{*}{nAP}}                 \\
			\midrule[0.9pt]
			  & 1-Shots &22.2&26.3&9.8\\
			  & 2-Shots &24.8&29.8&9.9 \\
			  & 3-Shots &26.1&31.5&10.1 \\
			  & 5-Shots &27.0&32.6&10.2 \\
			  \multirow{-5}{*}{TFA} & 10-Shots &27.9&33.9&10.0 \\
			\midrule[1pt]
			  & 1-Shots &28.8&34.9&\textbf{10.5}\\
			  & 2-Shots &30.0&36.5&\textbf{10.5} \\
			  & 3-Shots &30.3&37.0&10.3 \\
			  & 5-Shots &30.5&37.2&10.4 \\
			  \multirow{-5}{*}{CFA} & 10-Shots &\textbf{30.8}&\textbf{37.6}&\textbf{10.5} \\ 
			\bottomrule[1.1pt]
	\end{tabular}}
	\vspace{-0.18cm}
	\caption{Impact of variable number of base shots on the catastrophic forgetting of base classes. We compare CFA against TFA \cite{TFA}. The experiments are conducted on MS-COCO dataset given 10-shots of the novel categories. We report the results on the FC classifier based detector model \cite{FasterR-CNN}.}
 	\vspace{-1em}
	\label{tab:base-shots-ablation-study}
\end{table}

\textbf{Influence of the number of base shots.} In \cref{tab:base-shots-ablation-study}, we show the effect of using unbalanced datasets by finetuning with $K=1,2,3,5,10$ base shots and $10$ novel shots. Albeit both methods nearly attain their novel task performance, TFA indicates higher sensitivity towards a lower number of base shots. In  case of using one base shot, TFA exhibits a significant drop in bAP ($\sim22.4\%$). On the other hand, CFA is shown to be more robust to fewer base shots, experiencing a ($\sim7.2\%$) reduction, and thus achieving less forgetting with fewer base shots. Using only 3-shots, CFA reaches a similar performance to the 10-shot scenario with only $0.6$ points reduction in bAP. Based on the previous observations, CFA is more memory efficient where it can leverage fewer base shots to assist in achieving less forgetting.  
\begin{table}[t!] \centering
	\setlength{\tabcolsep}{3.0mm}
	\scalebox{0.8}
	{\begin{tabular}{c|ccc}
			\toprule[1.1pt]
			\multicolumn{1}{c|}{} &\multicolumn{3}{c}{}\\
			\multicolumn{1}{c|}{\multirow{-2}{*}{Model}}  & \multicolumn{1}{c}{\multirow{-2}{*}{AP}} & \multicolumn{1}{c}{\multirow{-2}{*}{bAP}}  &  \multicolumn{1}{c}{\multirow{-2}{*}{nAP}}            \\
			\midrule[0.9pt]
			  A-GEM w/ fc & 30.1 & 36.8 & 10.1\\
			  \rowcolor[HTML]{EFEFEF}
			  CFA w/ fc & 30.8 & \textbf{37.6} & 10.5 \\\midrule[1pt]
			  A-GEM w/ cos & 28.2 & 34.5 & 9.3  \\
			  \rowcolor[HTML]{EFEFEF}
			  CFA w/ cos & 30.3 & 36.6 & 11.3 \\\midrule[1pt]
			  A-GEM-DeFRCN & 30.3 & 35.6 & 14.4 \\
			  \rowcolor[HTML]{EFEFEF}
			  CFA-DeFRCN & \textbf{31.4} & 35.5 & \textbf{19.1}\\
			  
			\bottomrule[1.1pt]
	\end{tabular}}
	\vspace{-0.2cm}
	\caption{Comparison between fine-tuning different models with CFA and A-GEM \cite{agem}. The results are reported for MS-COCO under the 10-shot setting. }
 	\vspace{-1em}
	\label{tab:cl-ablation-study}
\end{table}

\textbf{Finetuning with A-GEM.} Since CFA extends A-GEM, we compare finetuning with A-GEM against CFA across the three models utilized throughout this work. As shown in \cref{tab:cl-ablation-study}, regardless of the model, CFA exhibits less forgetting and helps improve the novel task performance compared to A-GEM. Moreover, not only does CFA consistently attain better overall performance, but it also improves the nAP. This implies that the proposed constraint in CFA results in more forward knowledge transfer because, in each update step, both base and novel gradients adaptively contribute to the gradient update rule, which lead to minimize the expected risk on both tasks.

 \begin{figure}[t!]
 \centering
 \includegraphics[width=1.0\linewidth]{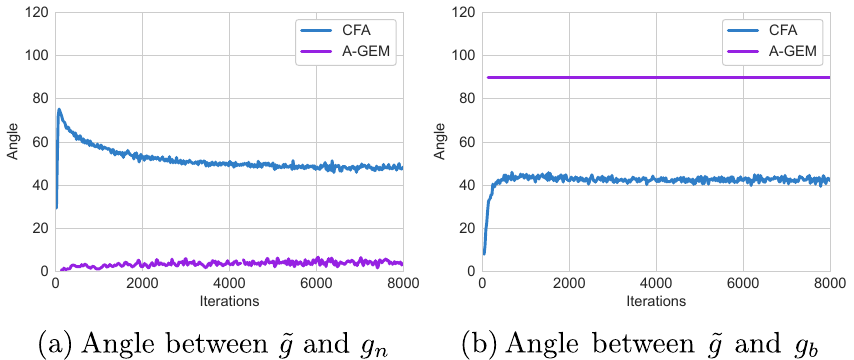}\vspace{-2mm}
 \caption{In 10-shot setting for MS-COCO dataset, the angle between the projected gradient $\tilde{g}$ and both $g_n$ (a) and $g_b$ (b) are depicted for A-GEM~\cite{agem} and CFA.\vspace{-1em}} %
 \label{fig:agem_cfa_comparisons}%
 \vspace{-2mm}
\end{figure}

\textbf{Impact of CFA and A-GEM on the projected gradient.} In~\cref{fig:agem_cfa_comparisons} (a), it is shown that as the network converges the projected gradient is quite close to the novel gradients direction ($\sim 4^\circ$) in case of A-GEM, whereas CFA provides a gradient update at a higher angle ($\sim 43^\circ$). This shows that A-GEM is biased towards the novel tasks compared to the base tasks. In~\cref{fig:agem_cfa_comparisons} (b), CFA provides a projected gradient much closer to the base task loss gradients ($\sim 45^\circ$) compared to the consistent orthogonal projection by the A-GEM. This has shown to allow for better transfer of knowledge when learning the novel tasks as well as achieving less forgetting.

\section{Conclusion}
We propose a new finetuning approach, CFA, for G-FSOD. It readapts a gradient episodic memory approach by replaying a few-shot of the base objects from a static memory buffer. A new gradient update rule is derived, which averages the base and novel gradients when the angle between them is acute. It also adaptively reweights them in case the novel gradients point towards a direction that could lead to forgetting. CFA strives to encourage the knowledge transfer between base and novel classes while being versatile enough to be integrated with FSOD frameworks without incurring overhead in model capacity or inference time. Experiments on MS-COCO and PASCAL-VOC demonstrate the effectiveness of the proposed method in G-FSOD on the overall AP. Moreover, CFA outperforms the state-of-the-art in FSOD on the novel AP. Ablation studies show that CFA is robust enough with a few base shots. In future work, we plan to extend our approach to a continual learning setting.    
\clearpage
{\small
\bibliographystyle{unsrt}
\bibliography{egbib}
}

\end{document}


\title{Supplementary Material}
\maketitle
\thispagestyle{empty}

 \begin{figure*}
 \centering
 \includegraphics[width=1.0\linewidth]{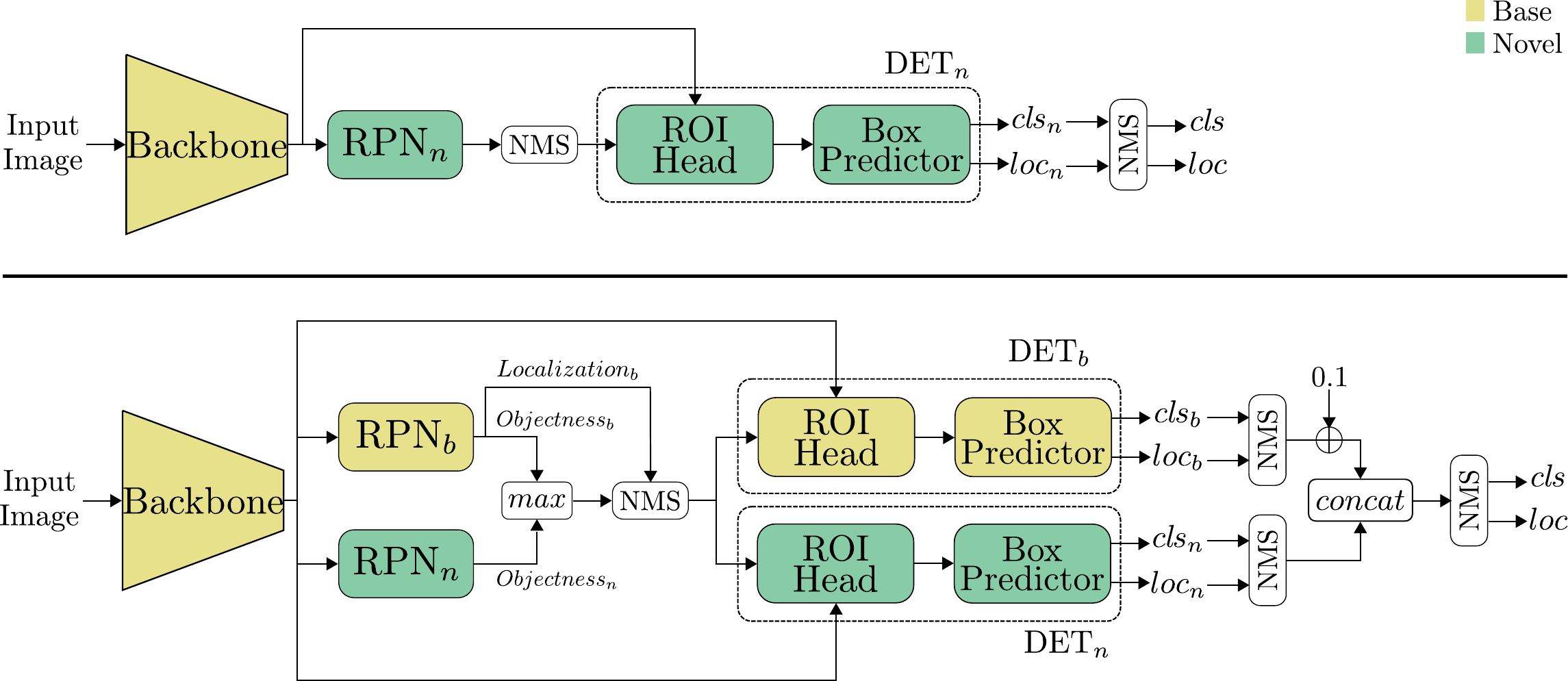} 
 \caption{\textbf{Top:} Illustration of the single model inference. \textbf{Bottom:} A detailed overview of the ensemble model evaluation protocol proposed by Retentive R-CNN \cite{gfsod}.} %
 \label{fig:ensemble_eval}
\end{figure*}
\section{Analytical Proof of CFA}
\label{sec:cfa_proof}
In this section, we mathematically derive the update rules for the proposed CFA method. We formulate our objective function as follows:
\begin{align}
	 	\textrm{minimize}_{\tilde{g}_n, \tilde{g}_b} \quad &\frac{1}{2}||g_n-\tilde{g}_n||_2^2 + \frac{1}{2}||g_b-\tilde{g}_b||_2^2 \nonumber\\
	 	\textrm{subject to} 				\quad &\tilde{g}_n^\top g_{b} \geq 0, \nonumber\\ 
	 	                                    \quad &\tilde{g}_b^\top g_{n} \geq 0,   \label{eq:qp}  
\end{align}
where $g_n$ and $g_b$ represents the proposed gradient update for the novel and base task, respectively. $\tilde{g}_n$ and $\tilde{g}_b$ denotes the projected gradient update for the novel and base task, respectively. If both constraints are satisfied, the update rule will be the average of $g_n$ and $g_b$. Otherwise, we solve the constrained optimization problem using the method of Lagrange multipliers.

First, we reformulate the problem in the standard form as follows:
\begin{align}
	 	\textrm{minimize}_{z_n, z_b} \quad &\frac{1}{2} z_n^{\top} z_n - g_n^{\top}z_n + \frac{1}{2} z_b^{\top} z_b - g_b^{\top}z_b\nonumber\\
	 	\textrm{subject to} 				\quad &-z_n^\top g_{b} \leq 0, \nonumber\\ 
	 	                                    \quad &-z_b^\top g_{n} \leq 0,   \label{eq:sqp}  
\end{align}
where $\tilde{g}_b$ and $\tilde{g}_n$ are denoted as $z_b$ and $z_n$, respectively. We ignore the constant terms $g_n^{\top}g_n$ and $g_b^{\top}g_b$. In addition, the sign of the inequality constraints is changed. Then, the Lagrangian can be formulated as:
\begin{align}
	\label{eq:lagrang}
	\mathcal{L}(z_n, z_b, \alpha_1, \alpha_2) = & \frac{1}{2}z_n^\top z_n - g_n^\top z_n - \alpha_1 z_n^\top g_b \nonumber \\
	& + \frac{1}{2}z_b^\top z_b - g_b^\top z_b - \alpha_2 z_b^\top g_n,
\end{align}
where $\alpha_1$ and $\alpha_2$ are the dual variables. To find the solution of the primal variables $z_n^*$ and $z_b^*$, we need to find the lower bound solution of the primal problem by computing the solution of the dual problem:
\begin{equation}
	\label{eq:dual}
	 \theta_{\mathcal{D}}(\alpha_1, \alpha_2) =  \min_{z_n, z_b} \mathcal{L}(z_n, z_b, \alpha_1, \alpha_2).
\end{equation}
 We find $z_n^*$ and $z_b^*$ as a function of dual variables $\alpha_1$ and $\alpha_1$, respectively, by minimizing the Lagrangian $\mathcal{L}(z_n, z_b, \alpha_1, \alpha_2)$. This is achieved by setting its derivatives w.r.t $z_n$ and $z_b$ to zero,
\begin{align}
	\nabla_{z_n} \mathcal{L}(z_n, z_b, \alpha_1, \alpha_2) = 0 \nonumber,\\
	z_n^* =   g_n + \alpha_1 g_{b},  \label{eq:zn_star}
\end{align}

\begin{align}
	\nabla_{z_b} \mathcal{L}(z_n, z_b, \alpha_1, \alpha_2) = 0 \nonumber,\\
	z_b^* =   g_b + \alpha_2 g_{n}.  \label{eq:zb_star}
\end{align}
Next, we can find the solution of the primal variables by solving the dual problem. We substitute \cref{eq:zn_star} and \cref{eq:zb_star} in \cref{eq:dual}. Now, the dual problem can be rewritten as:
\begin{align*}
	 \theta_{\mathcal{D}}(\alpha_1, \alpha_2) & = \frac{1}{2}(g_n^\top g_n + 2\alpha_1 g_n^\top g_{b} + \alpha_1^2 g_{b}^\top g_{b}) \\
	 & - g_n^\top g_n -2\alpha_1 g_n^\top g_{b} - \alpha_1^2 g_{b}^\top g_{b} \\
	 & + \frac{1}{2}(g_b^\top g_b + 2\alpha_2 g_b^\top g_{n} + \alpha_2^2 g_{n}^\top g_{n}) \\
	 & - g_b^\top g_b -2\alpha_2 g_b^\top g_{n} - \alpha_2^2 g_{n}^\top g_{n} \\
	 & =  -\frac{1}{2}g_n^\top g_n - \alpha_1 g_n^\top g_{b} - \frac{1}{2} \alpha_1^2 g_{b}^\top g_{b} \\
	 & -\frac{1}{2}g_b^\top g_b - \alpha_2 g_b^\top g_{n} - \frac{1}{2} \alpha_2^2 g_{n}^\top g_{n}.
\end{align*}
Next, we find the solution $\alpha_1^*$ and $\alpha_2^*$ of dual problem as follows:
\begin{align}
	\nabla_{\alpha_1} \theta_{\mathcal{D}}(\alpha_1, \alpha_2) = 0 \nonumber, \\
	\alpha_1^* = - \frac{g_n^\top g_{b}}{g_{b}^\top g_{b}},
	\label{eq:alpha1_star}
\end{align}
\begin{align}
	\nabla_{\alpha_2} \theta_{\mathcal{D}}(\alpha_1, \alpha_2) = 0 \nonumber, \\
	\alpha_2^* = - \frac{g_b^\top g_{n}}{g_{n}^\top g_{n}}.
	\label{eq:alpha2_star}
\end{align}
\begin{table*}[t!]
    \centering
    \begin{tabular}{c | c c c | c c c | c c c}
      \Xhline{1pt}
      \multirow{2}{*}{\textbf{Methods} / \textbf{Shots}} &
      \multicolumn{3}{c |}{\textbf{5 shot}}  &
      \multicolumn{3}{c |}{\textbf{10 shot}}  &
      \multicolumn{3}{c}{\textbf{30 shot}}\\
      & \textbf{AP} &  \textbf{AP50} & \textbf{AP75} & \textbf{AP} &  \textbf{AP50} & \textbf{AP75} &  \textbf{AP}  & \textbf{AP50} & \textbf{AP75} \\ \Xhline{1pt}
      
      FRCN-ft-full\;\cite{FasterR-CNN}\,$\ddag$\,$\S$ & 4.6 & 8.7 & 4.4 & 5.5 & 10.0 & 5.5 & 7.4 & 13.1 & 7.4 \\ 
      Meta-YOLO\;\cite{FSRW} &  - & - & - & 5.6 & 12.3 & 4.6 & 9.1 & 19.0 & 7.6 \\
      Meta R-CNN\;\cite{MetaRCNN} &  - & - & - & 8.7 & 19.1 & 6.6 & 12.4 & 25.3 & 10.8 \\
      TFA w/ cos\;\cite{TFA}\,$\ddag$\,$\S$ & 7.0 & 13.3 & 6.5 & 9.1 & 17.1 & 8.8 & 12.1 & 22.0 & 12.0 \\
      Meta\,Det\;\cite{MetaDet} &  - & - & - & 7.1 & 14.6 & 6.1 & 11.3 & 21.7 & 8.1 \\
      FSOD\;\cite{FSOD-RPN} &  - & - & - & 12.0 & 22.4 & 11.8 & - & - & - \\
      FsDetView\;\cite{FsDetView}\,$\S$ & 10.7 & 24.5 & 6.7 & 12.5 & 27.3 & 9.8 & 14.7 & 30.6 & 12.2 \\
      MPSR\;\cite{MPSR}\,$\ddag$ & 7.4 & 12.3 & 7.7 & 9.8 & 17.9 & 9.7 & 14.1 & 25.4 & 14.2  \\
      FSCE\;\cite{FSCE}\,$\ddag$\,$\S$ &  - & - & - & 11.1 & - & 9.8 & 15.3 & - & 14.2 \\
      CME\;\cite{CME}\,$\ddag$ &  - & - & - & 15.1 & 24.6 & 16.4 & 16.9 & 28.0 & 17.8 \\
      Deformable-DETR-ft-full\,\cite{DETR}\,$\S$ &  - & - & - & 11.7 & 19.6 & 12.1 & 16.3 & 27.2 & 16.7 \\
      DeFRCN\;\cite{defrcn} &15.5 & \textbf{29.4} & 14.2 & 18.3 & 33.7 & 17.4 & 22.6 & 39.8 & 22.8 \\
      \Xhline{1pt}
      \rowcolor[HTML]{EFEFEF}  
      CFA-DeFRCN (Ours) & \textbf{15.6} & 29.1 & \textbf{15.2} & \textbf{19.1} & \textbf{34.8} & \textbf{18.7} & \textbf{23.0} & \textbf{40.5} & \textbf{23.0} \\
    \Xhline{1pt}
    \end{tabular}
    \caption{Few-shot detection performance on MS-COCO for the novel categories.  $\ddag$ indicates methods using multi-scale features. $\S$ indicates results averaged on multiple runs.}
\label{tab:coco_novel}
\vspace{-1em}
\end{table*}

Given the solutions of the dual problem, we can find closed form solutions of $\tilde{g}_n$ and $\tilde{g}_b$ by substituting the dual solutions $\alpha_1^*$ \cref{eq:alpha1_star} and $\alpha_2^*$ \cref{eq:alpha2_star} in \cref{eq:zn_star} and \cref{eq:zb_star}, respectively:
\begin{equation}
\label{eq:gn_update_rule}
	z_n^* = g_n  - \frac{g_n^\top g_{b}}{g_{b}^\top g_{b}} g_{b} = \tilde{g}_n,
\end{equation}

\begin{equation}
\label{eq:gb_update_rule}
	z_b^* = g_b  - \frac{g_b^\top g_{n}}{g_{n}^\top g_{n}} g_{n} = \tilde{g}_b.
\end{equation}
After finding the closed form solution, a single update rule can be realized as:
\begin{equation}
    \tilde{g} = \frac{\tilde{g}_n + \tilde{g}_b}{2}.
\end{equation}

\addtolength{\tabcolsep}{-4.5pt}    

\setlength{\mywidth}{0.2\textwidth}

\bgroup
\def\arraystretch{0.5}
\begin{figure*}[t!]
\begin{tabular} {ccccc}

\includegraphics[width=\mywidth, height=\mywidth]{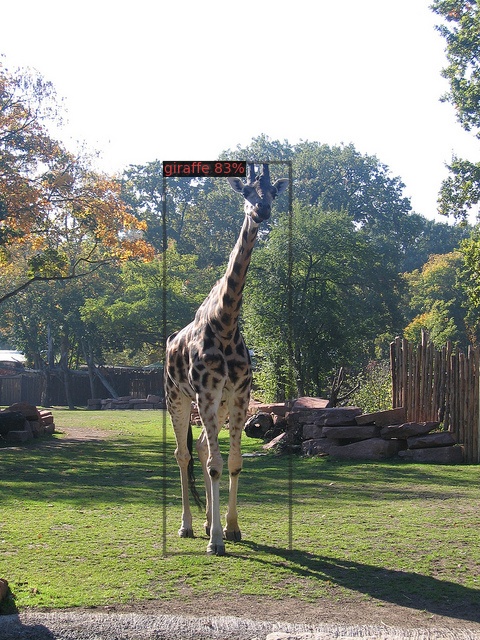} &  
\includegraphics[width=\mywidth, height=\mywidth]{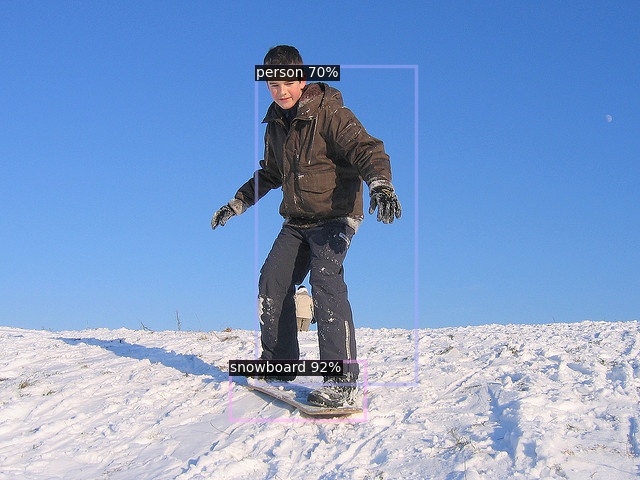} &   
\includegraphics[width=\mywidth, height=\mywidth]{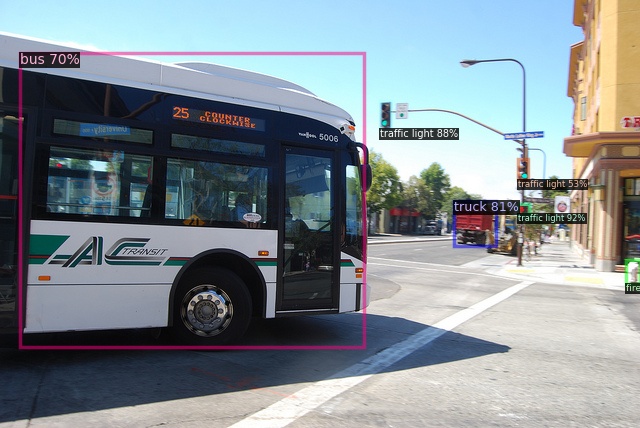} &   
\includegraphics[width=\mywidth, height=\mywidth]{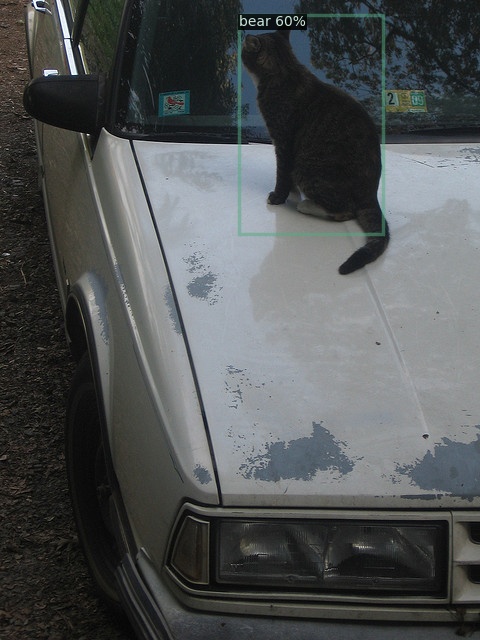} &
\includegraphics[width=\mywidth, height=\mywidth]{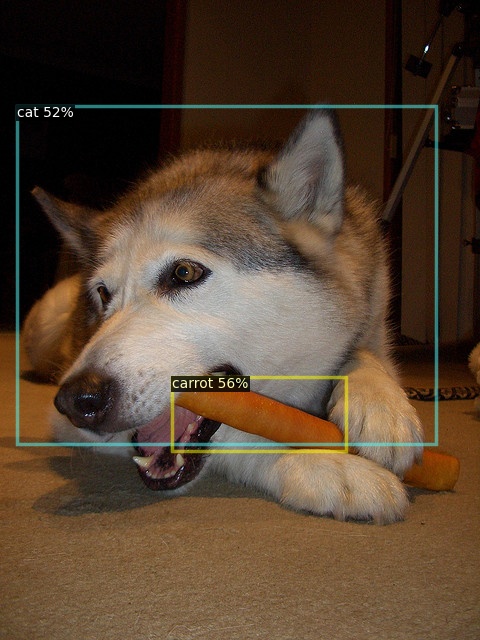} \\

\includegraphics[width=\mywidth, height=\mywidth]{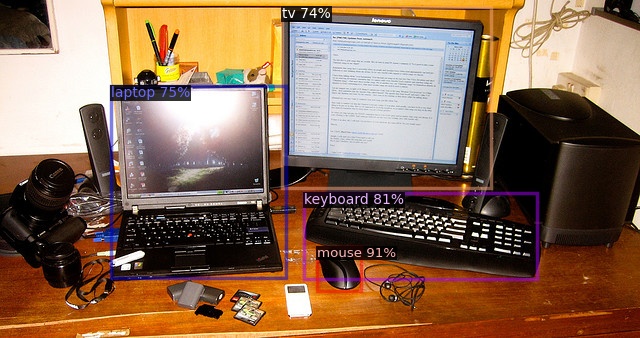} &  
\includegraphics[width=\mywidth, height=\mywidth]{images/success/COCO_val2014_000000009553.jpg} &   
\includegraphics[width=\mywidth, height=\mywidth]{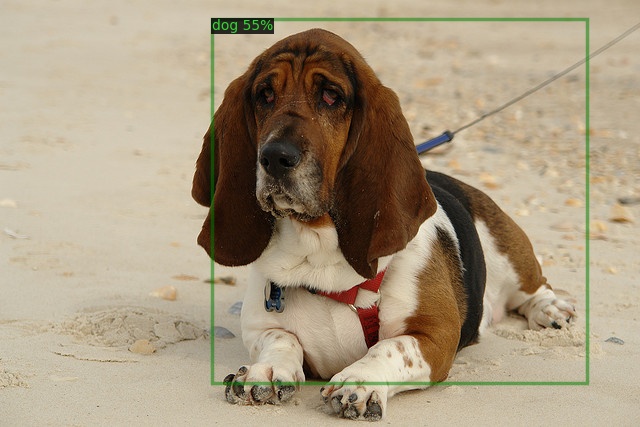} &  
\includegraphics[width=\mywidth, height=\mywidth]{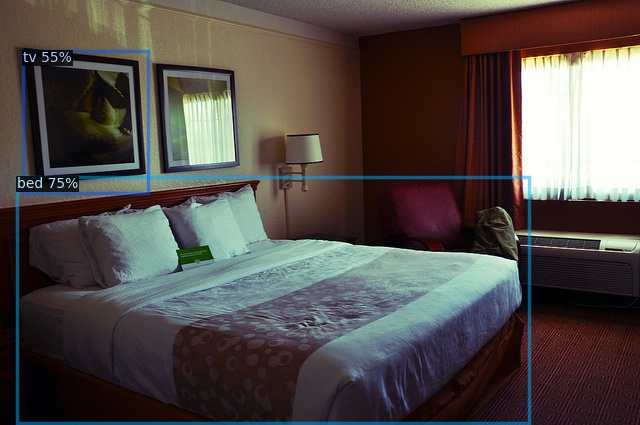} &   
\includegraphics[width=\mywidth, height=\mywidth]{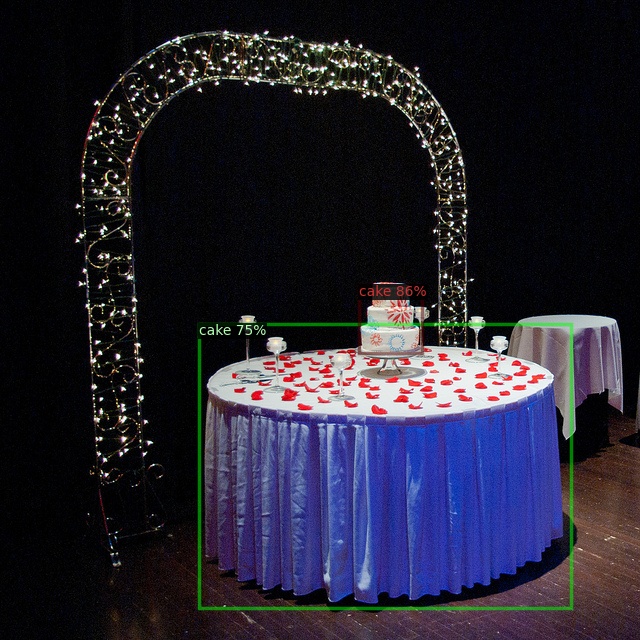} \\

\includegraphics[width=\mywidth, height=\mywidth]{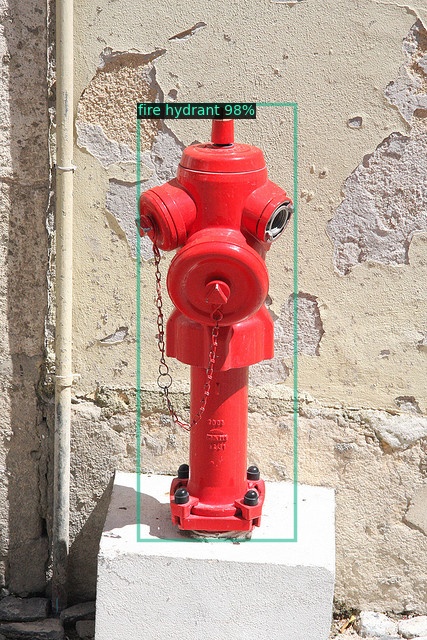} &  
\includegraphics[width=\mywidth, height=\mywidth]{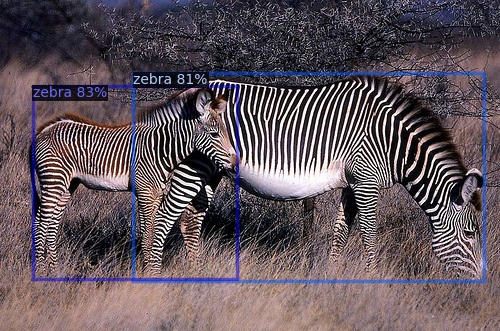} &   
\includegraphics[width=\mywidth, height=\mywidth]{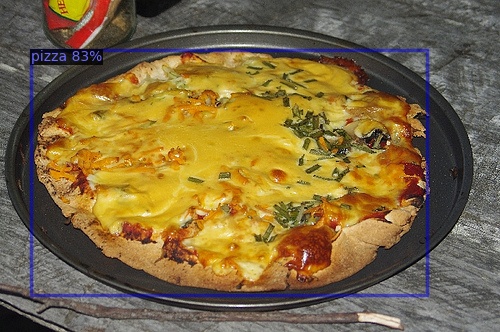} &  
\includegraphics[width=\mywidth, height=\mywidth]{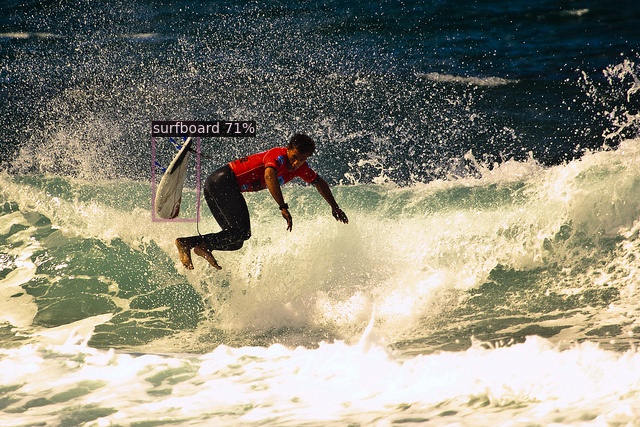} &   
\includegraphics[width=\mywidth, height=\mywidth]{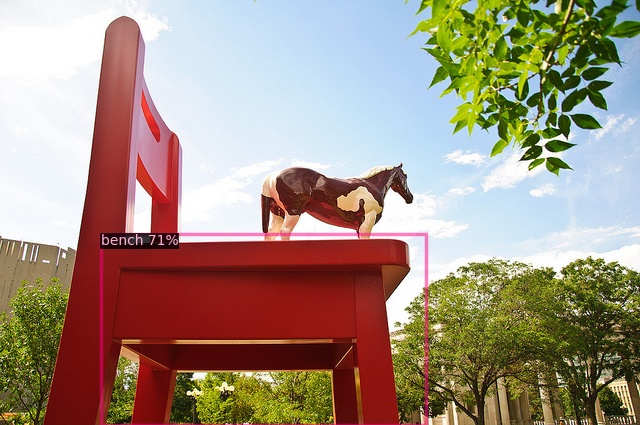} \\

\includegraphics[width=\mywidth, height=\mywidth]{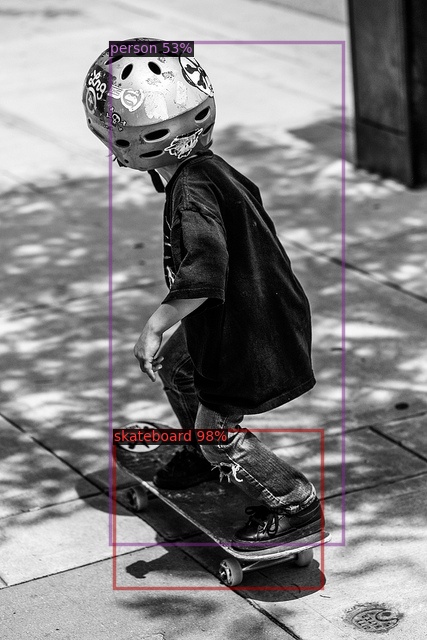} &  
\includegraphics[width=\mywidth, height=\mywidth]{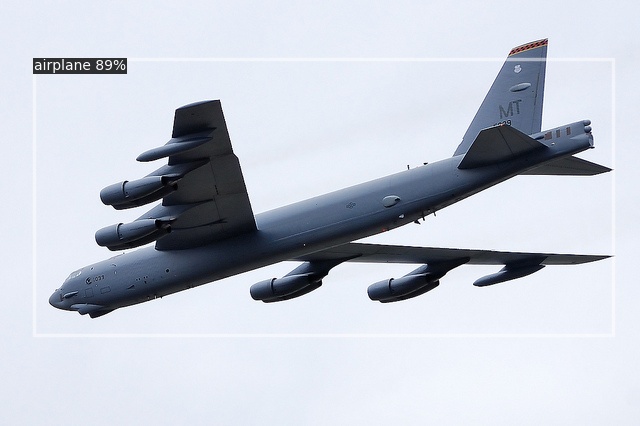} &
\includegraphics[width=\mywidth, height=\mywidth]{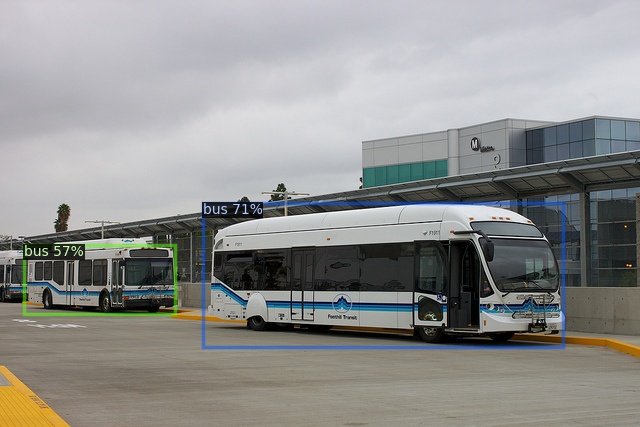} &   
\includegraphics[width=\mywidth, height=\mywidth]{images/failure/COCO_val2014_000000132375.jpg} &  
\includegraphics[width=\mywidth, height=\mywidth]{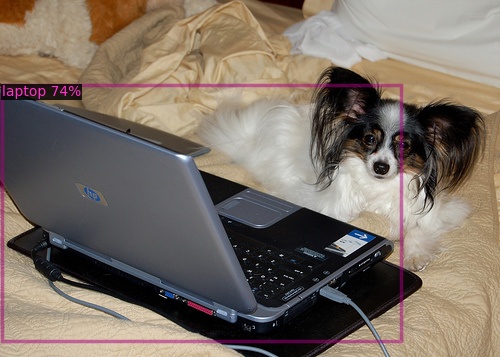} \\

\includegraphics[width=\mywidth, height=\mywidth]{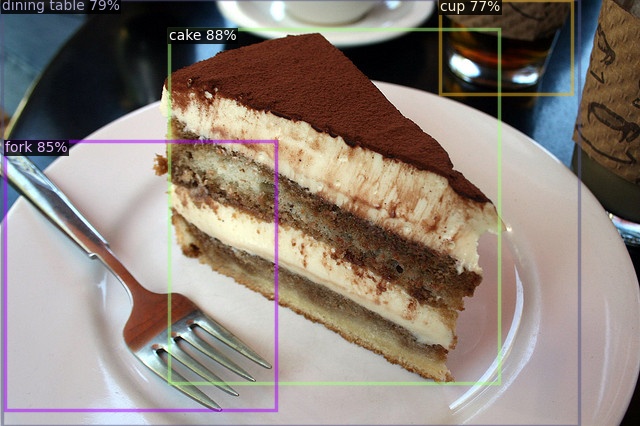} &  
\includegraphics[width=\mywidth, height=\mywidth]{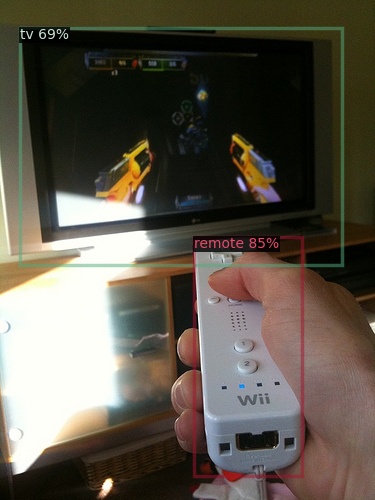} &
\includegraphics[width=\mywidth, height=\mywidth]{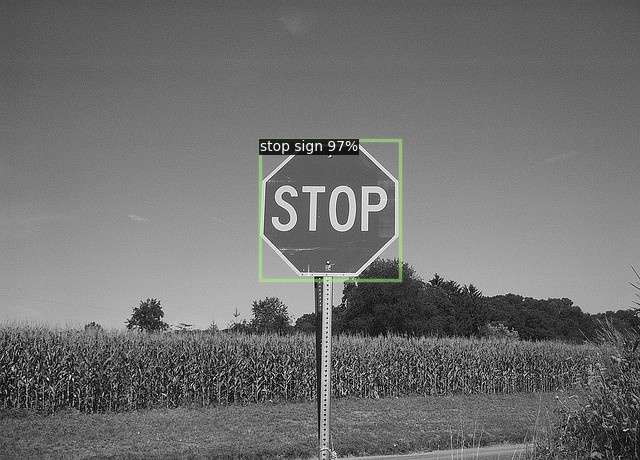} &   
\includegraphics[width=\mywidth, height=\mywidth]{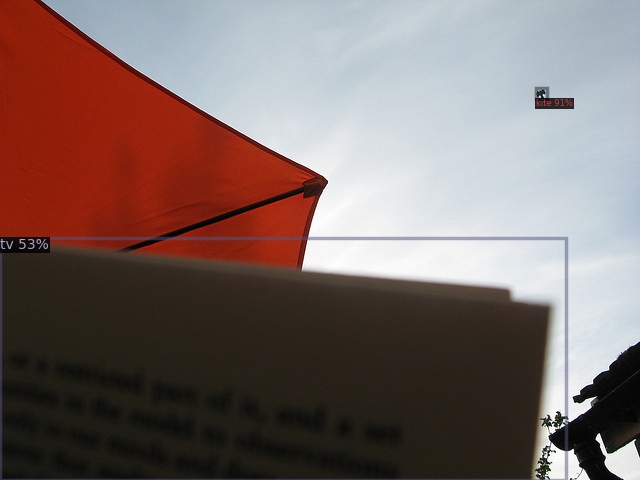} &  
\includegraphics[width=\mywidth, height=\mywidth]{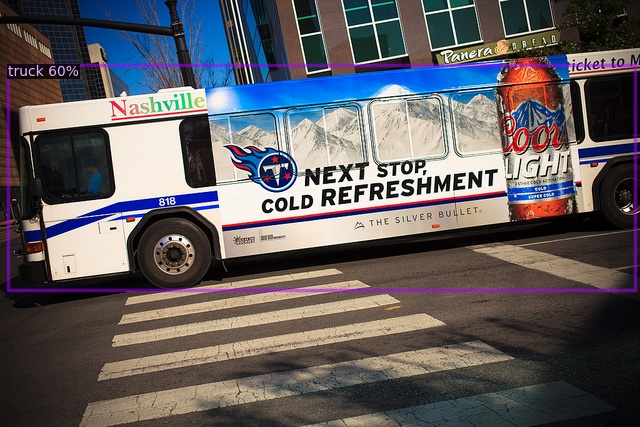} \\
 
\end{tabular}
	\caption{Qualitative analysis of the proposed CFA method on the MS-COCO dataset. The shown results are based on CFA w/cos finetuned under 30-shot setting. The first three columns show success scenarios while the last two columns present the failure scenarios.\vspace{-1em}}
	\label{fig::qualitative}
 \end{figure*}
 \egroup
 \addtolength{\tabcolsep}{4.5pt}

\definecolor{tfa_fc}{RGB}{0,95,115}
\definecolor{tfa_cos}{RGB}{0,58,82}
\definecolor{cfa_fc}{RGB}{255,208,0}
\definecolor{cfa_cos}{RGB}{202,103,2} 
\definecolor{defrcn}{RGB}{155,34,38}
\definecolor{cfa_defrcn}{RGB}{117,26,29}

\begin{figure*}
\centering
\begin{tikzpicture}
\begin{axis}[%
width=1.0\textwidth,
height=2.3in,
ybar=0pt,
bar width=0.1,
xtick={0,0.5,1.5,2.5,3.5,4.0},
xticklabels={{},5-shot,10-shot,30-shot,{}},
x tick label as interval,
enlargelimits={abs=0.5},
legend style={at={(0.5,-0.3)}, {nodes={scale=0.90}}, anchor=south,  legend columns=-1}
]

\addplot+[color=tfa_fc, fill=tfa_fc!70, draw=tfa_fc, error bars/.cd, 
y dir=both, y explicit]
coordinates {
    (1,25.6) +- (0.0, 0.5)
    (2,26.2) +- (0.0, 0.5)
    (3,28.4) +- (0.0, 0.3)};
\addplot+[color=tfa_cos, fill=tfa_cos!70, draw=tfa_cos, error bars/.cd,
y dir=both,y explicit]
coordinates {
    (1,25.9) +- (0.0, 0.6)
    (2,26.6) +- (0.0, 0.5)
    (3,28.7) +- (0.0, 0.4)};
\addplot+[color=cfa_fc, fill=cfa_fc!70, draw=cfa_fc, error bars/.cd,
y dir=both,y explicit]
coordinates {
    (1, 29.1) +- (0.0, 0.3)
    (2,29.9) +- (0.0, 0.3)
    (3,30.8) +- (0.0, 0.2)};
\addplot+[color=cfa_cos, fill=cfa_cos!70, draw=cfa_cos, error bars/.cd,
y dir=both,y explicit]
coordinates {
    (1,29.3) +- (0.0, 0.2)
    (2,30.2) +- (0.0, 0.2)
    (3,31.1) +- (0.0, 0.1)};
\addplot+[color=defrcn, fill=defrcn!70, draw=defrcn, error bars/.cd,
y dir=both,y explicit]
coordinates {
    (1,27.8) +- (0.0, 0.3)
    (2,29.7) +- (0.0, 0.2)
    (3,31.4) +- (0.0, 0.1)};
\addplot+[color=cfa_defrcn, fill=cfa_defrcn!70, draw=cfa_defrcn, error bars/.cd, y dir=both,y explicit]
coordinates {
    (1,28.4) +- (0.0, 0.2)
    (2,30.2) +- (0.0, 0.2)
    (3,31.7) +- (0.0, 0.1)};
\legend{TFA w/fc, TFA w/cos, CFA w/fc , CFA w/cos, DeFRCN, CFA-DeFRCN}
   
\end{axis}


\end{tikzpicture}%
\caption{Results over 10 random runs on MS-COCO under $K=5,10,30$-shot setting. The mean and $95\%$ confidence interval are reported.}
\label{fig:barchart}
\end{figure*}
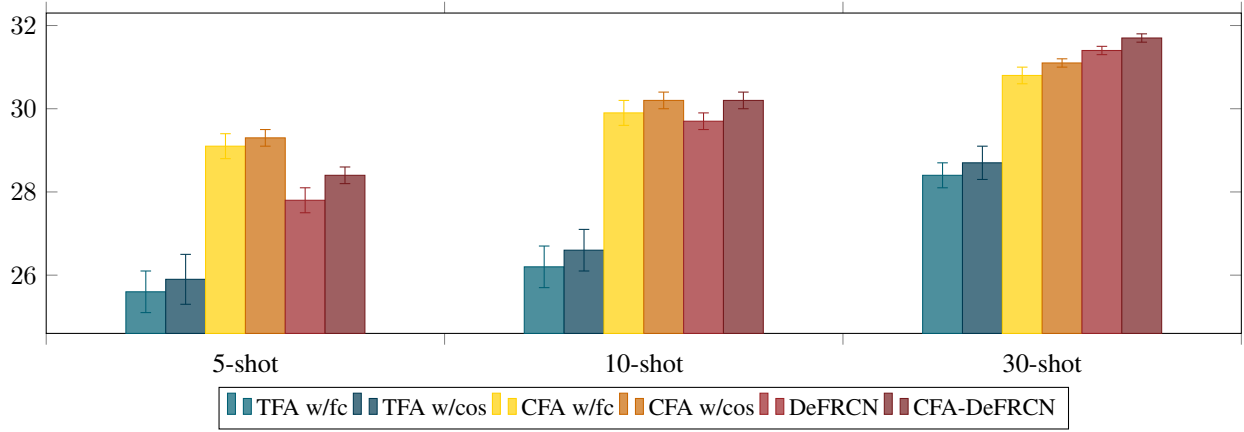
\begin{table*}
    \centering
    \scalebox{0.9}{
    \begin{tabular}{c | c c c | c c c | c c c}
      \Xhline{1pt}
      \multirow{2}{*}{\textbf{Methods} / \textbf{Shots}} & 
      \multicolumn{3}{c |}{\textbf{5 shot}}  &
      \multicolumn{3}{c |}{\textbf{10 shot}}  &
      \multicolumn{3}{c}{\textbf{30 shot}}\\
      & \textbf{AP} &  \textbf{bAP} & \textbf{nAP} & \textbf{AP} &  \textbf{bAP} & \textbf{nAP} & \textbf{AP} &  \textbf{bAP} & \textbf{nAP} \\ \Xhline{1pt}
      TFA w/ fc\cite{TFA} & 25.6$\pm$0.5 & 31.8$\pm$0.5 & 6.9$\pm$0.7 & 26.2$\pm$0.5 & 32.0$\pm$0.5 & 9.1$\pm$0.5 & 28.4$\pm$0.3 & 33.8$\pm$0.3 & 12.0$\pm$0.4 \\
      
      TFA w/ cos\cite{TFA} & 25.9$\pm$0.6 & 32.3$\pm$0.6 & 7.0$\pm$0.7 & 26.6$\pm$0.5 & 32.4$\pm$0.6 & 9.1$\pm$0.5 & 28.7$\pm$0.4 & 34.2$\pm$0.4 & 12.1$\pm$0.4 \\
      
      CFA w/ fc  & 29.1$\pm$0.3 & 36.2$\pm$0.3 & 7.7$\pm$0.6 & 29.9$\pm$0.3 & 36.7$\pm$0.2& 9.6$\pm$0.6 & 30.8$\pm$0.2 & 36.6$\pm$0.2 & 13.6$\pm$0.3\\
      
      CFA w/ cos  & 29.3$\pm$0.2 & 36.0$\pm$0.2 & 9.2$\pm$0.5& 30.2$\pm$0.2 & 36.6$\pm$0.1& 11.2$\pm$0.5 & 31.1$\pm$0.1 & 36.6$\pm$0.1 & 14.8$\pm$0.2\\
      
      DeFRCN\cite{defrcn} & 27.8$\pm$0.3 & 32.6$\pm$0.3 & 13.6$\pm$0.7 & 29.7$\pm$0.2 & 34.0$\pm$0.2 & 16.8$\pm$0.6 & 31.4$\pm$0.1 & 34.8$\pm$0.1 & 21.2$\pm$0.4\\
        
      CFA-DeFRCN & 28.4$\pm$0.2 & 32.8$\pm$0.2 & 15.2$\pm$0.5 & 30.2$\pm$0.2 & 34.0$\pm$0.2 & 18.8$\pm$0.4 & 31.7$\pm$0.1 & 34.6$\pm$0.1 & 23.0$\pm$0.3\\
      \hline
    \end{tabular}}
    \caption{G-FSOD experimental results for 5,10,30-shot settings on MS-COCO. We report AP, bAP, nAP for all, base, and novel classes, respectively. }
    \label{tab:coco}
    \vspace{-1em}
\end{table*}

\section{Evaluation Protocols}
\label{sec:eval_protocals}
In \cref{fig:ensemble_eval}, the utilized evaluation protocols are presented. The single model inference comprises the RPN$_n$ and DET$_n$, finetuned with a few-shots from the novel data, while the backbone is kept frozen. The evaluation is conducted as follows: (1) the image is fed to the backbone (2) the RPN$_n$ generates proposals (3) the proposals with IoU scores lower than a predefined threshold are omitted via a non-maximum suppression (NMS) (4) the DET$_n$ outputs both the classification logits $cls_n$ and bounding boxes $loc_n$, respectively (5) finally, the final predictions are filtered via a NMS. 

On the other hand, the ensemble inference model further employs the RPN$_b$ and DET$_b$ from the base model. The inference is done as follows: (1) the image is fed to the backbone (2) the image features are fed to both the RPN$_b$ and RPN$_n$ to compute the objectness logits $O_b$ and $O_n$, respectively (3) the maximum between $O_b$ and $O_n$ is fed to NMS along with the bounding boxes from RPN$_b$ (4) the filtered proposals are then fed to both  DET$_b$ and DET$_n$ to output the classification logits and bounding boxes (5) after the detectors' predictions are fed separately to a NMS,  a bonus of 0.1 are added to $cls_b$ (6) finally the output from both detectors are concatenated and fed to a NMS to output the final predictions. We emphasize that we did not use the ensemble models during finetuning (as in Retentive-RCNN \cite{gfsod}), but rather we finetuned a single model and used both the base and finetuned models during inference. 

\section{Qualitative Results}
In \cref{fig::qualitative}, we present qualitative results on CFA w/cos finetuned with 30-shot setting. The first three columns show various success scenarios while the last two columns show different failure cases. Compared to base classes, the model is less confident with novel categories. This can be attributed to learning indiscriminative features, hence resulting in false positives and false negatives.

\section{Additional Experiments}
\label{sec:mult_runs}
\textbf{Comparison against FSOD baselines.} To further investigate the impact of CFA on the novel classes, we compare the performace of CFA-finetuned models (CFA w/fc, CFA w/cos and CFA-DeFRCN) with FSOD models on the challenging MS-COCO benchmark. CFA-DeFRCN outperforms existing approaches on the novel AP metric, although it was trained in a G-FSOD setting which generally leads to lower performance on the novel classes. The results are shown in \cref{tab:coco_novel}.

\textbf{Multiple runs.} We run the CFA-finetuned models (CFA w/fc, CFA w/cos, and CFA-DeFRCN) using 10 different seeds on MS-COCO and compare with the baselines (TFA \cite{TFA} and DeFRCN \cite{gfsod}). The results are shown in \cref{tab:coco} and \cref{fig:barchart}. We use the same random seeds as TFA \cite{TFA} and DeFRCN \cite{defrcn}. CFA consistently improves the overall AP while displaying a narrower confidence interval.

\begin{table}[t!] \centering
	\setlength{\tabcolsep}{1.7mm}
	\scalebox{0.9}
	{\begin{tabular}{c|c|c|c}
			\toprule[1.1pt]
			\multicolumn{1}{c|}{}            & \multicolumn{1}{c|}{}                  &\multicolumn{1}{c|}{} &\multicolumn{1}{c}{} \\
			\multicolumn{1}{c|}{\multirow{-2}{*}{Method}}      &\multicolumn{1}{c|}{\multirow{-2}{*}{w/E}}                        & \multicolumn{1}{c|}{\multirow{-2}{*}{Inference Time (\si{\milli\second})}}            & \multicolumn{1}{c}{\multirow{-2}{*}{Model Capacity}} \\
			\midrule[0.9pt]
			  TFA w/ fc & \xmark &85&60.6M\\
			  TFA w/ cos & \xmark &87&60.6M \\
			  CFA w/ fc & \xmark &85&60.6M \\
			  CFA w/ cos & \xmark &86&60.6M \\
			  CFA-DeFRCN & \xmark &147&52.7M \\
			\midrule[1pt]
			  CFA w/ fc & \cmark &211&75.4M \\
			  CFA w/ cos & \cmark &211&75.4M \\
			  CFA-DeFRCN & \cmark &376&105.3M \\
			\bottomrule[1.1pt]
	\end{tabular}}
	\vspace{-0.18cm}
	\caption{Inference time and model capacity for different evaluation protocols. Ensemble methods have a significant overhead compared to single model. w/E denotes whether the ensemble method is employed.}
	\label{tab:base-shots-ablation-study}
\end{table}

\section{Further Ablation Experiments}
\label{sec:ablations}

\textbf{Inference time and model capacity.} We study the impact of the two evaluation methods (single model vs ensemble model) on the inference time and number of parameters during inference. Although ensemble model evaluation achieves less forgetting, the inference time increases in average by $52\%$. On the other hand, the number of parameters increases by $50\%$ in CFA w/fc (and w/cos) and by $102\%$ in CFA-DeFRCN, since DeFRCN unfreezes the backbone during finetuning.
\begin{table}[t!] 
    \centering
	\setlength{\tabcolsep}{1.7mm}
	\scalebox{0.86}
	{\begin{tabular}{c|ccc|ccc}
			\toprule[1.1pt]
			\multicolumn{1}{c|}{}            & \multicolumn{3}{c|}{}                  &\multicolumn{3}{c}{} \\
			\multicolumn{1}{c|}{\multirow{-2}{*}{Model}}      &\multicolumn{1}{c}{\multirow{-2}{*}{Backone}} & \multicolumn{1}{c}{\multirow{-2}{*}{RPN}}        & \multicolumn{1}{c|}{\multirow{-2}{*}{RoI Head}}                       & \multicolumn{1}{c}{\multirow{-2}{*}{AP}}            & \multicolumn{1}{c}{\multirow{-2}{*}{bAP}}  &  \multicolumn{1}{c}{\multirow{-2}{*}{nAP}}                 \\
			\midrule[0.9pt]
			  &&&& 27.9 & 33.9 & 10.0 \\
			  & &\cmark&&29.9&37.2& 7.9\\
			  &  &&\cmark&28.9&35.4&9.6 \\
			  &&\cmark&\cmark&28.9&35.1&10.2 \\
			\multirow{-5}{*}{TFA w/ fc}  &\cmark &\cmark&\cmark&24.1&29.0&9.1 \\
			\hline
			  & &&&29.6&36.0&10.4\\
			  & &\cmark&&30.3&37.4&9.3 \\
			  &  &&\cmark&\textbf{30.8}&\textbf{37.8}&9.6 \\
			  & &\cmark&\cmark&\textbf{30.8} & 37.6 & \textbf{10.5} \\
			\multirow{-5}{*}{CFA w/ fc}  &\cmark &\cmark&\cmark&23.9&28.6&10.1 \\
			\midrule[1.5pt]
			  &&&&28.7 & 35.0 & 10.0 \\
			  & &\cmark&&28.9&35.8&8.3 \\
			  &  &&\cmark&29.0&35.3&10.3 \\
			  & &\cmark&\cmark&29.2&35.2&11.2 \\
			 \multirow{-5}{*}{TFA w/ cos}  &\cmark &\cmark&\cmark&24.1&28.5&10.9 \\
			 
			\hline
			  &&&& 29.4&35.9&9.8\\
			  &&\cmark&&28.7&35.3&8.9 \\
			  &&&\cmark&30.2&\textbf{36.8}&10.6 \\
			  & &\cmark&\cmark&\textbf{30.3} & 36.6 & \textbf{11.3} \\
			\multirow{-5}{*}{CFA w/ cos}  &\cmark &\cmark&\cmark&23.6&27.9&10.9 \\

			\bottomrule[1.1pt]
	\end{tabular}}
	\vspace{-0.18cm}
	\caption{Effect of unfreezing different components of our detection model in comparison to TFA \cite{TFA}. \cmark\ denotes unfreezing a component. The results are reported for MS-COCO under 10-shots.}
	\label{tab:unfreezing_extended}
\end{table}
\begin{table}[t] \centering
	\setlength{\tabcolsep}{1.7mm}
	\scalebox{0.86}
	{\begin{tabular}{c|c|ccc}
			\toprule[1.1pt]
			\multicolumn{1}{c|}{}            & \multicolumn{1}{c|}{}                  &\multicolumn{3}{c}{} \\
			\multicolumn{1}{c|}{\multirow{-2}{*}{Model}}      &\multicolumn{1}{c|}{\multirow{-2}{*}{Base-Shots}}                        & \multicolumn{1}{c}{\multirow{-2}{*}{AP}}            & \multicolumn{1}{c}{\multirow{-2}{*}{bAP}}  &  \multicolumn{1}{c}{\multirow{-2}{*}{nAP}}                 \\
			\midrule[0.9pt]
			  & 1-Shots &22.2&26.3&9.8\\
			  & 2-Shots &24.8&29.8&9.9 \\
			  & 3-Shots &26.1&31.5&10.1 \\
			  & 5-Shots &27.0&32.6&10.2 \\
			  \multirow{-5}{*}{TFA w/ fc} & 10-Shots &27.9&33.9&10.0 \\
			\hline
			  & 1-Shots &28.8&34.9&\textbf{10.5}\\
			  & 2-Shots &30.0&36.5&\textbf{10.5} \\
			  & 3-Shots &30.3&37.0&10.3 \\
			  & 5-Shots &30.5&37.2&10.4 \\
			  \multirow{-5}{*}{CFA w/ fc} & 10-Shots &\textbf{30.8}&\textbf{37.6}&\textbf{10.5} \\ 
			 \midrule[1.5pt]
			  & 1-Shots &24.2&28.9&10.0\\
			  & 2-Shots &26.5&32.0&10.2 \\
			  & 3-Shots &27.2&32.9&10.3 \\
			  & 5-Shots &27.8&33.6&10.3 \\
			  \multirow{-5}{*}{TFA w/ cos} & 10-Shots &28.7&35.0&10.0 \\
			\hline
			  & 1-Shots &28.6&34.3&\textbf{11.5}\\
			  & 2-Shots &29.8&35.9&11.3 \\
			  & 3-Shots &30.0&36.2&11.3 \\
			  & 5-Shots &30.2&36.4&11.3 \\
			  \multirow{-4}{*}{CFA w/ cos} & 10-Shots &\textbf{30.3}&\textbf{36.6}&11.3 \\
			\bottomrule[1.1pt]
	\end{tabular}}
	\vspace{-0.18cm}
	\caption{Impact of variable number of base shots on the catastrophic forgetting of base classes. We compare CFA against TFA \cite{TFA}. The experiments are conducted on MS-COCO dataset given 10-shots of the novel categories.}
	\vspace{-0.3cm}
	\label{tab:base_shots_extended}
\end{table}

\textbf{Extended ablation study.} We extend the ablation studies conducted on TFA w/fc and CFA w/fc to include the TFA w/cos and CFA w/cos. The results are presented in \cref{tab:unfreezing_extended} and \cref{tab:base_shots_extended}.

{\small
\bibliographystyle{unsrt}
\bibliography{egbib}
}